\RequirePackage{fix-cm}
\documentclass[twocolumn]{svjour3}          % twocolumn
\smartqed  % flush right qed marks, e.g. at end of proof
\usepackage{graphicx}

\usepackage{times}
\usepackage{epsfig}
\usepackage{graphicx}
\usepackage{amsmath}
\usepackage{amssymb}
\usepackage{bm}
\usepackage{color}
\usepackage{subfigure}
\usepackage{url}
\usepackage{natbib}

\newcommand{\argmin}{\mathop{\mathrm{argmin}}}

\begin{document}

\title{WhittleSearch: Interactive Image Search with Relative Attribute Feedback
%Image Search with Relative Attribute Feedback and Attribute Pivots
%\thanks{Grants or other notes
%about the article that should go on the front page should be
%placed here. General acknowledgments should be placed at the end of the article.}
}
%\subtitle{Do you have a subtitle?\\ If so, write it here}

%\titlerunning{Short form of title}        % if too long for running head

\author{Adriana Kovashka \and
        Devi Parikh \and
        Kristen Grauman
}

%\authorrunning{Short form of author list} % if too long for running head

\institute{Adriana Kovashka \at
University of Pittsburgh\\
5325 Sennott Square, 210 South Bouquet Street, Pittsburgh, PA, 15260, USA\\
\email{kovashka@cs.pitt.edu}\\
(Work done while at The University of Texas at Austin.) % \\ 
              \and Kristen Grauman \at
              The University of Texas at Austin\\
              2317 Speedway, Stop D9500, Austin, TX 78712\\
              \email{grauman@cs.utexas.edu}          % \\
           \and
           Devi Parikh \at
           Virginia Tech\\
           1185 Perry St, Room 302, MC 0111, Blacksburg, VA 24061\\
           \email{parikh@vt.edu}
}

\date{Received: date / Accepted: date}
% The correct dates will be entered by the editor

\maketitle

\begin{abstract}
We propose a novel mode of feedback for image search, where a user describes which properties of exemplar images should be adjusted in order to more closely match his/her mental model of the image sought.  For example, perusing image results for a query ``black shoes'', the user might state, ``Show me shoe images like these, but \emph{sportier}.''  Offline, our approach first learns a set of ranking functions, each of which predicts the relative strength of a nameable attribute in an image (e.g., \emph{sportiness}).  At query time, the system presents the user with a set of exemplar images, and the user relates them to his/her target image with comparative statements.  Using a series of such constraints in the multi-dimensional attribute space, our method iteratively updates its relevance function and re-ranks the database of images.  To determine which exemplar images receive feedback from the user, we present two variants of the approach: one where the feedback is user-initiated and another where the feedback is actively system-initiated.  In either case, our approach allows a user to efficiently ``whittle away'' irrelevant portions of the visual feature space, using semantic language to precisely communicate her preferences to the system.  We demonstrate our technique for refining image search for people, products, and scenes, and we show that it outperforms traditional binary relevance feedback in terms of search speed and accuracy.  In addition, the ordinal nature of relative attributes helps make our active approach efficient---both computationally for the machine when selecting the reference images, and for the user by requiring less user interaction than conventional passive and active methods.

%Insert your abstract here. Include keywords, PACS and mathematical
%subject classification numbers as needed.
% \PACS{PACS code1 \and PACS code2 \and more}
% \subclass{MSC code1 \and MSC code2 \and more}
\end{abstract}

%%%%%%%%% BODY TEXT
\section{Introduction}

In image search, the user often has a mental picture of his or her desired content.   For example, a shopper wants to retrieve those catalog pages that match his envisioned style of clothing; a witness wants to help law enforcement locate a suspect in a database based on his memory of the face; a web page designer wants to find a stock photo suitable for her customer's brand image.  Therefore, a central challenge is how to allow the user to convey that mental picture to the system.  Due to the well known ``semantic gap"---which separates the system's low-level image representation from the user's high-level concept---retrieval through a single user interaction, i.e., a one-shot query, is generally insufficient.  Keywords alone are clearly not enough; even if all existing images were tagged to enable keyword search, it is infeasible to pre-assign tags sufficient to satisfy any future query a user may dream up.  Indeed, vision algorithms are necessary to further parse the \emph{content} of images for many search tasks.  Advances in image descriptors, learning algorithms, and large-scale indexing have all had impact in recent years.

The key to overcoming the gap appears to be \emph{interactive} search techniques that allow a user to iteratively refine the results retrieved by the system~\citep{Cox00,Kurita93,Rui98,Zhou03,Ferecatu07,Zavesky08}.  The basic idea is to show the user candidate results, obtain feedback, and adapt the system's relevance ranking function accordingly.  However, existing image search methods provide only a narrow channel of feedback to the system.  Typically, a user refines the retrieved images via binary feedback on exemplars deemed ``relevant'' or ``irrelevant''~\citep{Kurita93,Cox00,Rui98,Zhou03,Ferecatu07}, or else attempts to tune system parameters such as weights on a small set of low-level features (e.g., texture, color, edges)~\citep{Flickner95,Ma97,Iqbal02}.  The latter is clearly a burden for a user who likely cannot understand the inner workings of the algorithm.  The former feedback is more natural to supply, yet it leaves the system to infer \emph{what about those images} the user found relevant or irrelevant, and therefore can be slow to converge on the user's target in practice.  The semantic gap between low-level visual cues and the high-level intent of a user remains, making it difficult for people to predict the behavior of content-based search systems.

\begin{figure}[t]
\includegraphics[width=1\linewidth]{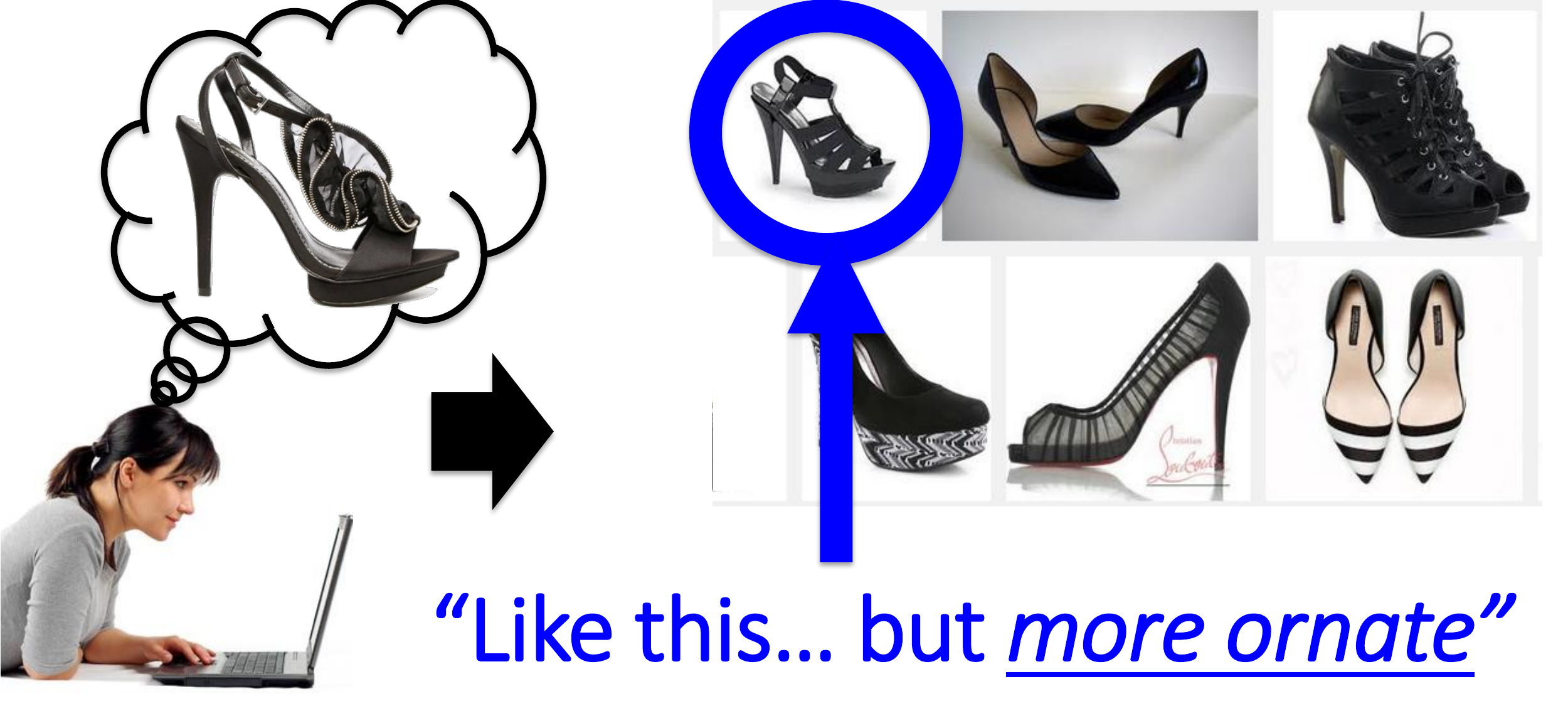}
\caption{Main idea: Allow users to give relative attribute feedback on reference images to refine their image search.}
\label{fig:concept-WS}
\end{figure}

In light of these shortcomings, we propose a novel mode of feedback where a user directly describes how high-level properties of exemplar images should be adjusted in order to more closely match his/her envisioned target images.  For example, when conducting a query on a shopping website, the user might state: ``I want shoes like these, but \emph{more formal}."  When browsing images of mug shots of suspects, a witness to a crime could say: ``He looked like this, but with \emph{longer hair} and a \emph{broader noise}."  When searching for stock photos to fit an ad, he might say: ``I need a scene \emph{similarly bright} as this one and \emph{more urban} than that one."  See Figure~\ref{fig:concept-WS}.  In this way, rather than simply state which images are (ir)relevant, the user employs semantic terms to say \emph{how} they are so.  Such feedback enables the system to more closely match the user's mental model of the desired content, with less total interaction effort compared to conventional click-based relevance feedback.  We call the approach \emph{WhittleSearch}, since it allows users to ``whittle away'' irrelevant portions of the visual feature space via precise, intuitive statements of their attribute preferences.

Briefly, our relative attribute feedback approach works as follows.  Offline, we first learn a set of ranking functions, each of which predicts the relative strength of a nameable attribute in an image (e.g., the degree of \emph{shininess}, \emph{furriness}, etc.).  At query time, the system presents some reference exemplar image(s), and the user provides relative attribute feedback on one or more of those images.  Using the resulting constraints in the multi-dimensional attribute space, we update the system's relevance function, re-rank the pool of images, and display to the user the next exemplar image(s).  This procedure iterates using the accumulated constraints until the top ranked images are acceptably close to the user's target.

In this pipeline, a key question is which exemplar images should be shown to the user for feedback.   To address this question, we explore two variants of the proposed WhittleSearch approach: one where the user decides which images require relative attribute feedback, and one where the system decides for which images it would most like the user's feedback.

\begin{figure*}[t]
\begin{tabular}{cc}
\subfigure[User-initiated feedback]{
\includegraphics[width=0.45\linewidth]{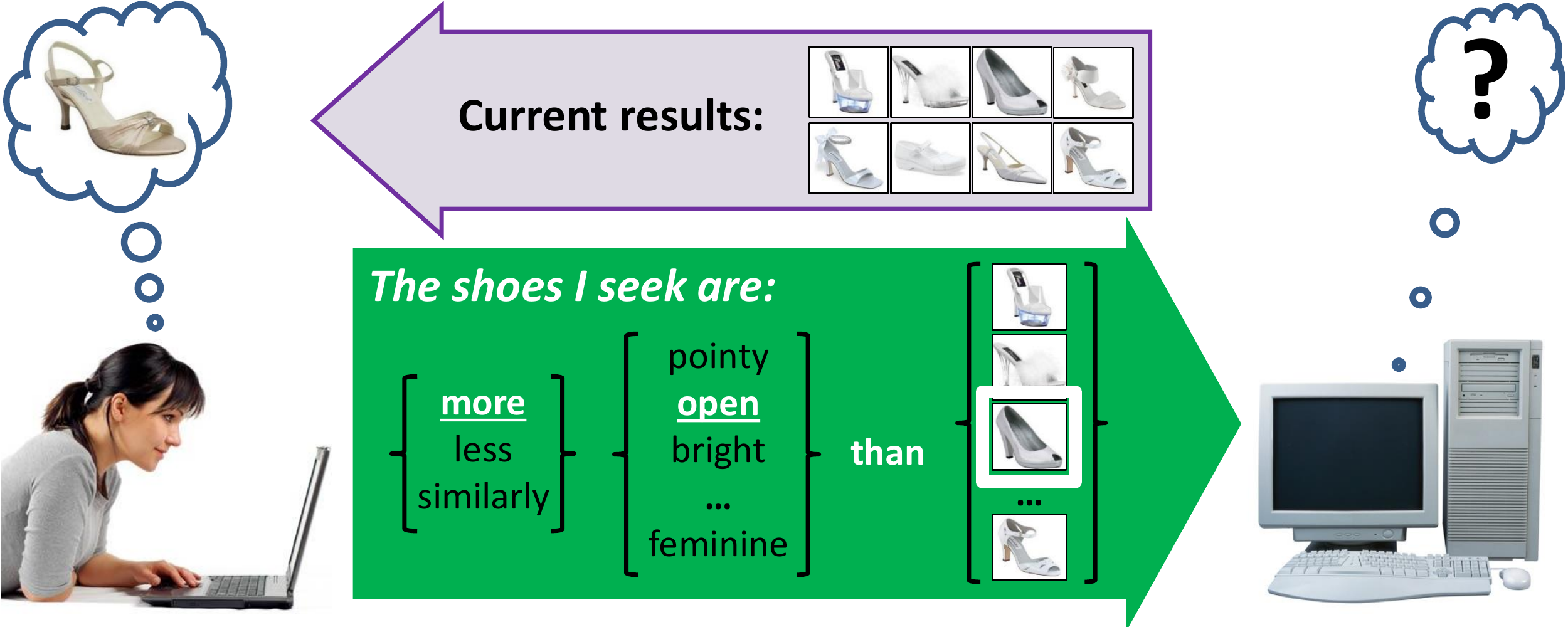}}
\hspace*{0.2in}
\subfigure[System-elicited feedback]{
\includegraphics[width=0.45\linewidth]{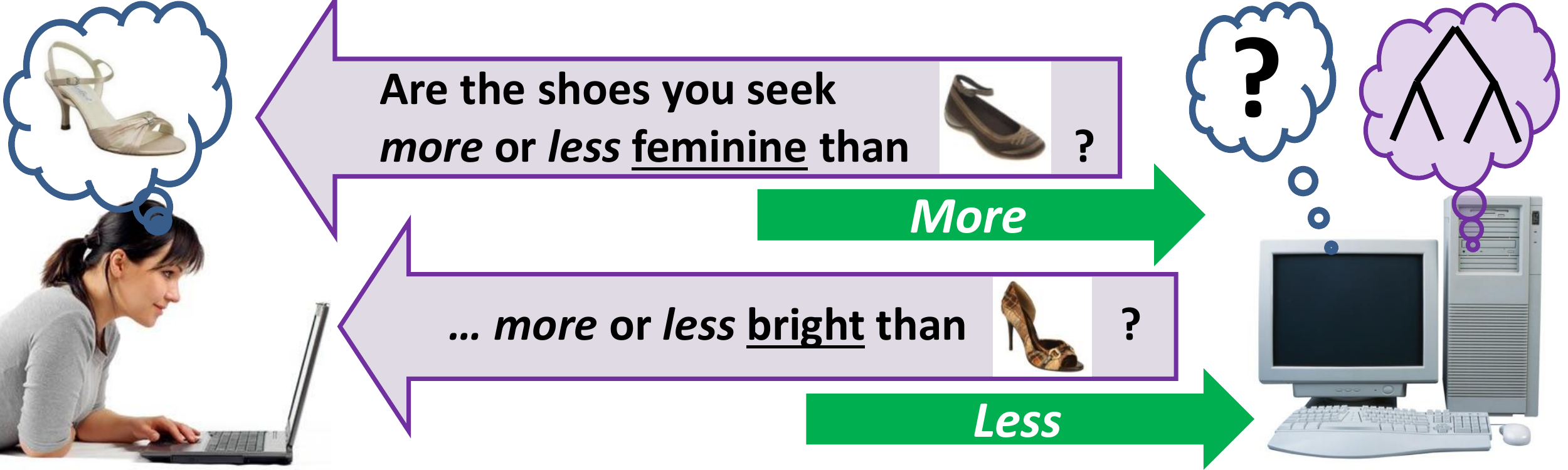}}
\end{tabular}
\caption{We consider two ways to elicit feedback for WhittleSearch: (a) a user-initiated approach, and (b) a system-initiated approach.  In (a), the user browses the current top-ranked images and decides what to comment on.  In (b), the system actively requests feedback on a specific image and attribute that is expected to most reduce its uncertainty about the relevance of the database images for this particular user query.}
\label{fig:concept-AWS}
\end{figure*}

In standard search interfaces, the user is shown a page of image results, i.e., those images the system currently estimates to be most relevant, and is free to react to any of them.  Similarly, in the first of the two WhittleSearch variants, we present the user with reference images consisting of the top-ranked most relevant images and allow him/her to generate feedback that pairs any of those images with any attribute in our vocabulary.  This setup gives the user the freedom to comment on exactly what he/she finds important for achieving good image results.  See Figure~\ref{fig:concept-AWS}(a).  Since the presented reference images are those currently ranked best by the system, this formulation has the additional advantage that the user is shown only those results that are increasingly similar to the target image.

However, the images believed to be most \emph{relevant} need not be most \emph{informative} for reducing the system's uncertainty.  Therefore, in the second WhittleSearch variant, we develop an \emph{active} approach for selecting the reference images for feedback. Intuitively, we want to solicit feedback on those exemplars that would most improve the system's notion of relevance.  Existing methods for actively guiding user feedback typically exploit classifier uncertainty to find useful exemplars, e.g.,~\citep{Tong01,Li01,Cox00,Zhou03}, or use clustering to distribute feedback among representative exemplars~\citep{Ferecatu07}.  Such traditional approaches have two main limitations.  First, the imprecision of binary relevance feedback (``Image X is relevant; image Y is not.") clouds the active selection criterion because extrapolation of the feedback to other images is unreliable.  Second, existing active selection techniques add substantial computational overhead to the interactive search loop, since ideally they must scan all database images to find the most informative exemplars.

Taking these shortcomings into account, in our Active WhittleSearch formulation, we propose to guide the user through a coarse-to-fine search using relative attributes.  At each iteration of feedback, the user provides a visual comparison between the attribute in his envisioned target and a ``pivot" exemplar, where a pivot separates all remaining relevant images into two balanced sets.  We show how to actively determine along which of multiple such attributes the user's comparison should next be requested, based on the expected information gain that would result.   The resulting algorithm is reminiscent of the popular 20-questions game---except the questions generated by the system are \emph{comparative} in nature.  See Figure~\ref{fig:concept-AWS}(b).

The active variant of our method works as follows.  Given a database of images, we first construct a binary search tree for each relative attribute of interest (e.g., \emph{pointiness}, \emph{shininess}, etc.).  Initially, the pivot exemplar for each attribute is the database image with the median relative attribute value.  Starting at the roots of these trees, we predict the information gain that would result from asking the user how his target image compares to each of the current pivots.  To compute the expected gain, we devise methods to estimate the likelihood of the user's response given the feedback history.  Then, among the pivots, the most informative comparison is requested, generating a question to the user such as, ``Is your target image more or less (or equally) pointy than this image?"  Following the user's response, the system updates its relevance predictions on all images and moves the current pivot down one level within the selected attribute's tree, unless the response is ``equally'', in which case we no longer need to explore this attribute tree.  

Notably, whereas prior information-gain methods would require a naive scan through all database images for each iteration, the proposed attribute search trees allow us to limit the scan to just one image per attribute.  Thus, our method is efficient both for the system (which analyzes a small number of candidates per iteration) and the user (who locates his content via a small number of well-chosen interactions).

Our main contribution is to widen human-machine communication for interactive image search by allowing users to communicate their preferences precisely and efficiently through \emph{visual comparisons}.  We demonstrate the two versions of WhittleSearch applied to several realistic search tasks for shoes, people, and scenes.  We compare our relative attribute feedback against traditional binary relevance feedback, and we show that it refines search results more effectively, often with less total user interaction.  We also present an approach which unifies the complementary strengths of relative attribute and binary feedback, allowing feedback of both types.  We quantify the advantages of the active selection of reference images over conventional active methods and a simpler binary search tree baseline that lacks our information gain prediction model.  The results strongly support our pivot-based approach as an efficient means to guide user feedback.

\section{Related Work}

\subsection{Interactive feedback in image search}

Relevance feedback has long been used to improve interactive image search~\citep{Kurita93,Cox00,Rui98,Tieu00,Ferecatu07,Zhou03}.  The main idea is to tailor the system's ranking function to the current user, based on his (usually iterative) feedback on the relevance of selected exemplar images.  This injects subjectivity into the model, implicitly guiding the search engine to pay attention to certain low-level visual cues more than others.  

In a binary relevance feedback model, the user identifies a set of relevant images and a set of irrelevant images among the current reference set.  
The user can also identify which images are \emph{more relevant} than others \citep{Ferecatu07}. While this is a relative comparison, just like in other binary relevance feedback methods, the system is not told \emph{in what way} image X is more relevant than image Y.
Image search results are produced by ranking all database images using a classifier (or some other statistical model), and the binary feedback supplies additional positive and negative training examples to enhance that classifier. 

Like existing interactive methods, our approach aims to elicit a specific user's target visual concept.  However, while prior work restricts input to the form ``A is relevant, B is not'' or, as suggested by \cite{Ferecatu07}, ``C is more relevant than D'', our approach allows users to comment precisely on what is missing from the current set of results.  We show that this richer form of feedback can lead to more effective refinement.  Being able to pinpoint \emph{how} one image is more relevant than another (via attributes) is the key contribution of our approach.

In practice, the images displayed to the user for feedback are usually those ranked best by the system's current relevance model.  However, if a user is cooperative, it can be more valuable to present a mix of probable relevant and irrelevant examples for feedback. If feedback is binary, with the user labeling examples as relevant (positive) or irrelevant (negative), the selection can naturally be cast as an active learning problem: the best examples to show are those that the relevance classifier is most uncertain about~\citep{MacArthur00,Tong01,Li01,Zhou03}.  Since focusing only on uncertain examples may ignore parts of the feature space, an alternative strategy is to display images representative of clusters in the database~\citep{Ferecatu07}.

Notably, prior efforts to display the exemplar image set that minimizes uncertainty were forced to resort to sampling or clustering heuristics due to the combinatorial optimization problem inherent when categorical feedback is assumed, e.g., \citep{Cox00,Ferecatu07}.  In contrast, we show that eliciting \emph{comparative} feedback on ordinal visual attributes naturally leads to an efficient sequential selection strategy, where each comparison is guaranteed to decrease the predicted relevance of half of the unexplored database images.

\subsection{Active testing and ``20 questions" labeling}

Whereas we are interested in actively eliciting user feedback during search, active methods are also relevant for choosing a series of useful ``tests" (e.g., features to extract) or label requests (``does the bird have a yellow beak?") for recognition tasks \citep{Geman98,Sznitman10,Vijayanarasimhan10,Branson10}.  
In the case where a \emph{human} answers the tests, attributes are well-suited to query for intermediate labels that will lead to the right high-level label, as demonstrated for bird labeling tasks~\citep{Branson10}.
Under certain scenarios, a globally optimal classification tree can be devised, so that an image is efficiently classified via a series of binary tests~\citep{Geman98}.  
Object localization problems also permit sequential search strategies that intelligently gather evidence within the image~\citep{Sznitman10,Vijayanarasimhan10}.  
A recent approach to categorization uses a human in the loop to provide responses to actively chosen similarity comparisons \citep{Wah14}.  
While this work employs relative comparisons, the problem setting is different than the one considered here.  That work performs categorization of an image provided to the system, not retrieval of images that match a user's mental model.

Our Active WhittleSearch idea shares the spirit of rapidly reducing uncertainty through a sequence of useful questions.  However, our aim is distinct.  Active testing entails selecting queries to classify a single novel image efficiently, i.e., reduce uncertainty over class labels for that image, whereas we select queries to efficiently find a target in a collection of images, i.e., reduce relevance uncertainty for all database images.  Moreover, our approach solicits visual \emph{comparisons}---key to eliminating irrelevant content in search---whereas prior work solicits traditional image labels.

\subsection{Attributes for image search}

Visual attributes are semantic properties of objects (e.g., \emph{fuzzy}, \emph{plastic}) that serve as a middle ground between low-level features (e.g., color, texture) and high-level categories.  When used in image search, the idea is to learn classifiers to predict the presence of various high-level semantic concepts from a lexicon---such as objects, locations, activity types, or properties---and then perform retrieval in the space of those predicted concepts.  Human-nameable \emph{semantic concepts} or attributes are often used in the multimedia community to build intermediate representations for image retrieval~\citep{Smith03,Rasiwasia07,Naphade06,Zavesky08,Douze11,WangX11,Scheirer12,WangX11,Douze11}.  They are especially valuable since they permit content-based keyword queries~\citep{Kumar08,Siddiquie11,Scheirer12,Rastegari13}.  While originally treated as categorical (``is \emph{smiling}'' vs.~``is not \emph{smiling}''), attributes can more generally be modeled as continuous or \emph{relative} properties (``is \emph{smiling more} than $X$'')~\citep{Parikh11b}.  While prior work demonstrates that attributes can provide a richer representation than raw low-level image features for image search, no previous work considers attributes as a handle for user feedback, as we propose.  In addition, we generalize the class-based training procedure used in~\citep{Parikh11b} to learn relative attributes, instead exploiting human-generated relative comparisons between image exemplars.

This manuscript unifies and expands the work we initially presented in~\citep{Kovashka12} and \citep{Kovashka13_pivots}, where we first proposed to use relative attributes as a feedback mechanism for image search.  
In this manuscript, we bring together the core approach of those two papers.  We analyze and discuss the advantages and disadvantages of the two forms of feedback, i.e., user-initiated free-form feedback and system-initiated active selection.  We  perform new experimental comparisons of the two versions of our method and examine when one is better than the other.  Finally, we introduce several new qualitative results.

\subsection{Attributes for recognition}

Apart from image search, attributes have also gathered interest in the object recognition community~\citep{Lampert09,Farhadi09,Kumar09,WangY10,Branson10,Patterson12,Wah13,Saleh13,Kulkarni14}.  Since attributes are often shared among object categories (e.g., \emph{made of wood}, \emph{plastic}, \emph{has wheels}), they are amenable to a number of interesting tasks, such as zero-shot learning from category descriptions~\citep{Lampert09,Parikh11b,Patterson12,Jayaraman14}, describing unfamiliar or anomalous objects~\citep{Farhadi09,Saleh13}, or categorizing with a 20-questions game \citep{Branson10}. We explore relative attributes in the distinct context of feedback for image search.

Other work investigates training object recognition classifiers with actively selected attribute labels.  By modeling object-attribute~\citep{Kovashka11,Parkash12,Biswas13} or attribute-attribute relationships~\citep{Zhang02,Mensink11}, one can request the most useful labels to refine the classifiers or propagate labels.  Our goal is quite different: we do active exemplar selection for image search, not classification, and our approach requests visual comparisons, not attribute labels.

\section{Approach}

Our approach allows a user to iteratively refine the search using feedback on attributes.  The user has some \emph{target} image in mind---the imagined visual content the user wants to locate in the database.  The target could be a literal image he/she has seen before, or simply a coarse mental model of the content of interest.  The user initializes the search with some keywords---either the name of the general class of interest (``shoes'') or some multi-attribute query (``black high-heeled shoes'')---and our system's job is to help refine from there.  If no such initialization is possible, we simply begin with a random set of images for feedback.  The top-ranked images are then displayed to the user, and the feedback-refinement loop begins.

Each iteration of the loop consists of the following: (a) a choice on the part of the system regarding which reference image(s) to the display to the user for feedback; (b) a choice on the part of the user regarding which reference image(s) to comment on and/or a decision about the relationship between the user's target and the reference image(s); and (c) an update of the system's notion of relevance, and thus the ranking of all images in the database.  

Throughout, let $\mathcal{D} = \{I_1,\dots,I_N\}$ refer to the pool of $N$ database images that are ranked by the system using its current scoring function $S_t : I \rightarrow \mathbb{R}$, where $t$ denotes the iteration of refinement.  The scoring function is trained using all accumulated feedback from iterations $1,\dots,t-1$, and it supplies an ordering (possibly partial) on the images in $\mathcal{D}$.  At each iteration, the top $K < N$ ranked images $\mathcal{T}_{t} = \{I_{t1},\dots,I_{tK}\} \subseteq \mathcal{D}$ are displayed to the user, where $S_t(I_{t1}) \geq S_t(I_{t2}) \geq \dots \geq S_t(I_{tK})$.   
A user then gives feedback of his choosing on any or all of the $K$ refined results in $\mathcal{T}_t$ (in the user-initiated WhittleSearch variant), or else he gives feedback specifically requested by the system on a particular image not necessarily among those in $\mathcal{T}_t$ (in the system-initiated WhittleSearch variant).

In the following, we first discuss how to learn the relative strength of an attribute in an image (Section \ref{sec:learn-attributes}).  Then we introduce the proposed new mode of relative attribute feedback and explain how the image search system uses this feedback to update its notion of relevance (Section \ref{sec:relevance}).  We then extend the idea to accommodate both our new relative attribute feedback and traditional binary feedback in a hybrid approach (Section \ref{sec:hybrid}).  Finally, we propose an approach to relegate to the system the choice of the reference images for feedback, and explain how to select the optimal reference image in each round of interaction (Section \ref{sec:active}).

\subsection{Learning to Predict Relative Attributes}
\label{sec:learn-attributes}

Suppose we have a vocabulary of $M$ attributes $\{a_m\}_{m=1}^M$, which may be generic or domain-specific for the image search problem of interest.  For example, a domain-specific vocabulary for shoe shopping could contain attributes such as \emph{shininess}, \emph{heel height}, \emph{colorfulness}, etc., whereas for scene descriptions it could contain attributes like \emph{openness}, \emph{naturalness}, \emph{depth}.  While we assume this vocabulary is given, recent work suggests it may also be  discovered automatically or semi-automatically~\citep{Berg10,Parikh11a,Maji12,Patterson12}.

Typically semantic visual attributes are learned as categories: a given image either exhibits the concept or it does not, and so a classification approach to predict attribute presence is sufficient~\citep{Rasiwasia07,Naphade06,Zavesky08,Lampert09,Farhadi09,Kumar09,WangY10,Douze11}.  In contrast, to express feedback in the form sketched above, we require \emph{relative} attribute models that can predict the degree to which an attribute is present.  Therefore, we first learn a \emph{ranking function} for each attribute in the given vocabulary.  Note that one might informally treat classifier outputs as ``strengths'', yet doing so is inconsistent with a training procedure that actually targets hard categorical labels.  Results in~\citep{Parikh11b} confirm that simply treating a binary classifier output value as the strength of presence is inferior in practice compared to training ranking functions.

\begin{figure}[t]
\centering
\includegraphics[width=1\linewidth]{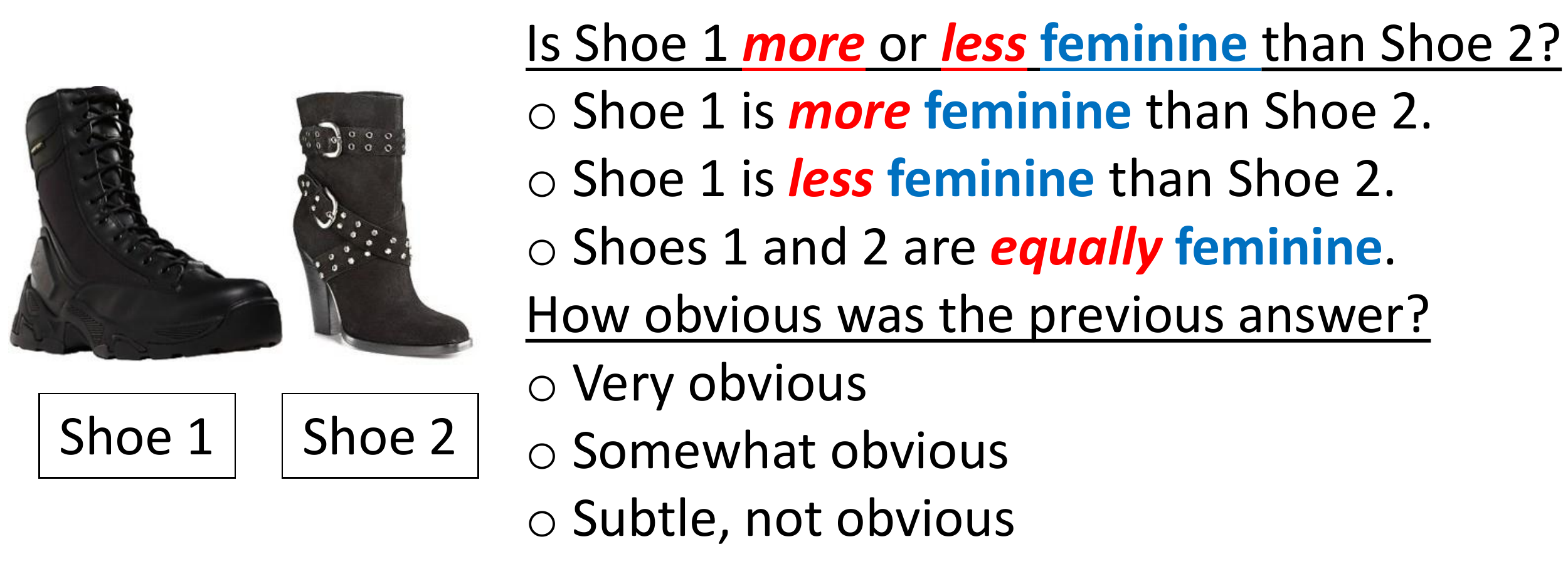}
\caption{Interface for image-level relative attribute annotations.}
\label{fig:relative-supervision}
\end{figure}

For each attribute $a_m$, we obtain supervision on a set of image pairs $(i,j)$ in a training set $\mathcal{I}$. We ask human annotators to judge whether that attribute has stronger presence in image $i$ or $j$, or if it is equally strong in both.  Such judgments can be subtle, so on each pair we collect up to five redundant responses from multiple annotators on Amazon Mechanical Turk (MTurk); see Figure \ref{fig:relative-supervision}.  To distill reliable relative constraints for training, we use only those for which most labelers agree.  This yields a set of ordered image pairs $O_m = \{(i,j)\}$ and a set of un-ordered pairs $E_m = \{(i,j)\}$ such that $(i,j) \in O_m \implies i \succ j$, i.e. image $i$ has stronger presence of attribute $a_m$ than image $j$, and $(i,j) \in E_m \implies i \sim j$, i.e. $i$ and $j$ have equivalent strengths of $a_m$.

We would like to emphasize the design for constraint collection: rather than ask annotators to give an \emph{absolute} score reflecting how much the attribute $m$ is present, we instead ask them to make \emph{comparative} judgements on two exemplars at a time.  This is both more natural for an individual annotator, and it also permits seamless integration of the supervision from many annotators, each of whom may have a different internal ``calibration'' for the attribute strengths.

Next, to learn an attribute's ranking function, we employ the large-margin formulation of Joachims~\citep{Joachims02}, which was originally shown for ranking web pages based on clickthrough data, and used for relative attribute learning~\citep{Parikh11b}.  Suppose each image $I_i$ is represented in $\mathbb{R}^d$ by a feature vector $\bm{x}_i$ (we use color and GIST; more details below).  We aim to learn $M$ ranking functions, one per attribute:
\begin{equation}
a_m(I_i) = \bm{w}_m^T \bm{x}_i,
\end{equation}
for $m=1,\dots,M$, such that the maximum number of the following constraints is satisfied:
\begin{align}
\forall (i,j) \in O_m: \bm{w}_m^T\bm{x}_i > \bm{w}_m^T\bm{x}_j. 
\end{align}

Joachims' algorithm approximates this NP-hard problem by introducing (1) a regularization term that prefers a wide margin between the ranks assigned to the closest pair of training instances, and (2) slack variables 
$\xi_{ij}$ on the constraints, yielding the following objective~\citep{Joachims02}:
\begin{align}\label{eq:joachimeq}
\textrm{minimize} & \quad \left(\frac{1}{2}||\bm{w_m}^T||_2^2+C\sum\xi_{ij}\right) \\
s.t. & \quad \bm{w}_m^T\bm{x}_i \geq \bm{w}_m^T\bm{x}_j +1 -\xi_{ij};~~~\forall (i,j) \in O_m \nonumber\\
     & \quad \xi_{ij} \geq 0,\nonumber
\end{align}
where $C$ is a constant penalty.  The objective is reminiscent of standard SVM training (and is solvable using similar decomposition algorithms), except the linear constraints enforce relative orderings rather than labels.  While shown here in the linear form, the method is also kernelizable.  We use Joachims' SVMRank code \citep{Joachims06}.\footnote{Note that one can also use the equality constraints in $E_m$ for training these ranking functions, as in \cite{Parikh11b}.  In our approach, we use these constraints to compute parameters for scoring relevance, in Section \ref{sec:relevance}.}

Having trained $M$ such functions, we are then equipped to predict the extent to which each attribute is present in any novel image, by applying the learned functions $a_1,\dots,a_M$ to its image descriptor $\bm{x}$.  This training is a one-time process done before any search query or feedback is issued.  Furthermore, the data $\mathcal{I}$ used for training attribute rankers is not to be confused with our database pool $\mathcal{D}$; the two may be disjoint sets of images.

Whereas \citet{Parikh11b} propose generating supervision for relative attributes from top-down category comparisons (``person X is (always) \emph{more smiley} than person Y''), our approach extends the learning process to incorporate \emph{image-level} relative comparisons (``image A exhibits \emph{more smiling} than image B'').  While training from category-level comparisons is clearly more expedient, we find that image-level supervision is important in order to reliably capture those attributes that do not closely follow category boundaries.  The \emph{smiling} attribute is a good example of this contrast, since a given person (the category) need not be smiling to an equal degree in each of his/her photos.  In fact, our user studies on MTurk show that category-level relationships violate 23\% of the image-level relationships specified by human subjects for the \emph{smiling} attribute.  In the results section, we detail related human studies analyzing the benefits of image-level comparisons.

\subsection{Relative Attribute Feedback}
\label{sec:relevance}

Next we define the basic WhittleSearch framework.  With the ranking functions learned above, we can now map any image from the database $\mathcal{D}$ into an $M$-dimensional space, where each dimension corresponds to the relative rank prediction for one attribute.  It is in this feature space we propose to handle query refinement from a user's feedback.

A user of the system has a mental model of the target visual content he seeks.  To refine the current search results, he surveys the $K$ top-ranked images in $\mathcal{T}_t$, and uses some of them as reference images with which to better express his envisioned optimal result.  These constraints are of the form ``What I want is more/less/similarly $A$ than image $I_{t_f}$", where $A$ is an attribute name, and $I_{t_f}$ is an image in $\mathcal{T}_t$ (the subscript $t_f$ denotes it is a reference image for $f$eedback at iteration $t$).   For now, suppose these relative constraints are given for some combination of image(s) and attribute(s) of the user's choosing.  Later, in Section~\ref{sec:active}, we will consider how instead the system can guide the choice of the image and attribute for feedback so as to most quickly reduce its uncertainty about what the user wants.

\begin{figure}[t]
\includegraphics[width=1\linewidth]{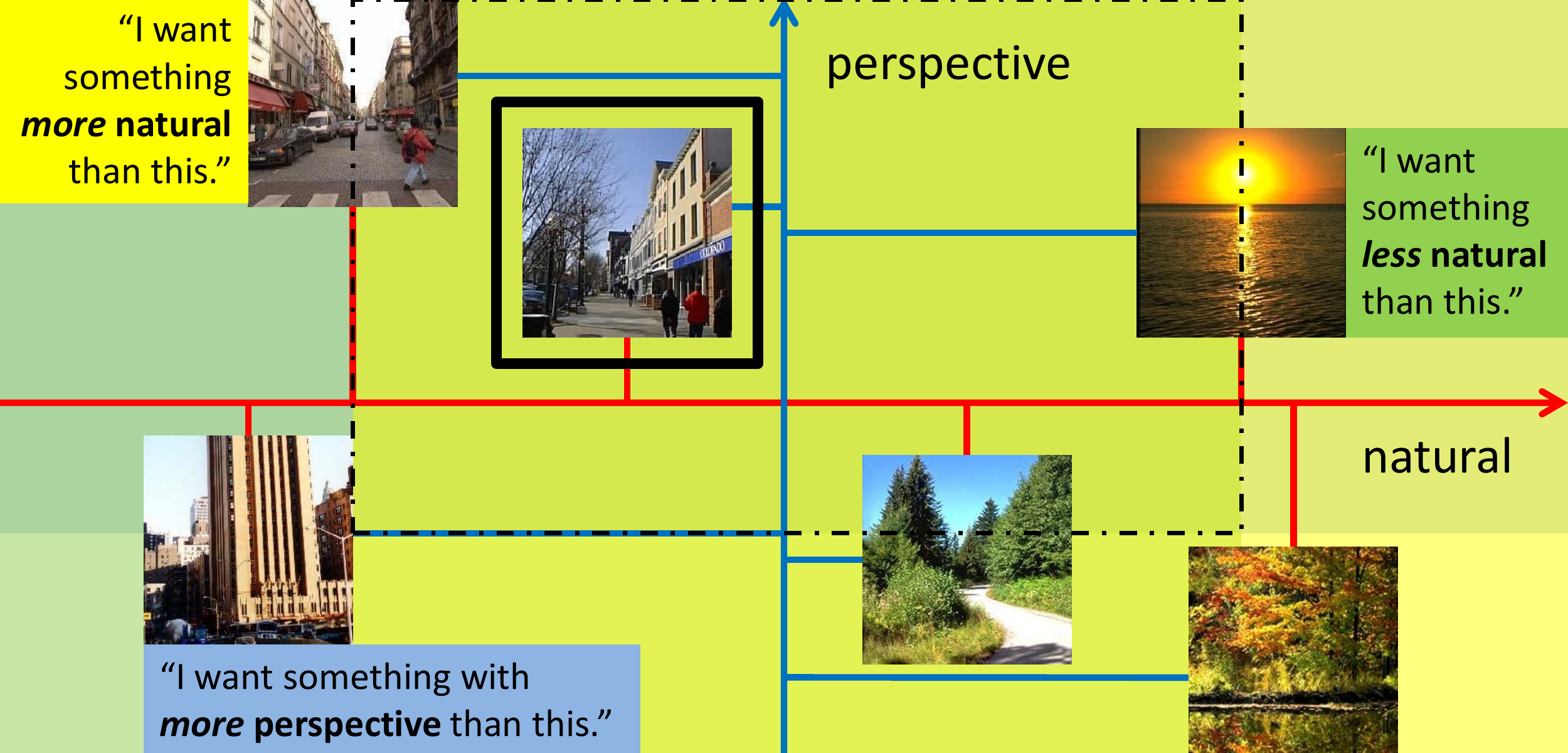}
\caption{Sketch of WhittleSearch relevance computation. This toy example illustrates the intersection of relative constraints with $M=2$ attributes. The images are plotted on the axes for both attributes. The space of images that satisfy each constraint are marked in a different color. The region satisfying all constraints is marked with a black dashed line. In this case, there is only one image in it (outlined in black). Best viewed in color.}
\label{fig:approach-WS}
\vspace{-0.1in}
\end{figure}

The WhittleSearch system accumulates this feedback \linebreak from the user during each round of interaction, each time updating the relevance it associates with each database image.  Intuitively, the user's statements about relative preferences serve to carve out a relevant region of the $M$-dimensional attribute feature space, whittling away images not meeting the user's requirements.  See Figure~\ref{fig:approach-WS}.
Accordingly, we next define a relevance function that predicts the extent to which a database image matches the user's target.  It is a probabilistic model of relevance to account for the fact that predicted attribute values can deviate from true perceived attribute strengths to some extent.\footnote{We do, however, assume that all users would agree on the true attribute strength in a given image.  See~\cite{Kovashka13_adapt} for an approach to model the user-specific perception of an attribute.}

Let $y_i \in \{1, 0\}$ denote the binary label for image $I_i$, which reflects whether it is relevant to the user (matches his target), or not.  Let $\mathcal{F}=\{I_{t_f}, m, r\}_{t=1}^T$ denote the set of comparative constraints accumulated in the $T$ rounds of feedback so far. The $t$-th item in $\mathcal{F}$, $\mathcal{F}_t$, consists of a reference image $I_{t_f}$ for attribute $m$, and a user response $r \in \{\text{``more''}, \text{``less''}, \text{``equally''}\}$.  The final output of our search system will be a sorting of the database images in $\mathcal{D}$ according to their probability of relevance, given the image content and all user feedback: $P(y_i = 1 | I_i, \mathcal{F})$, for $i=1,\dots,N$.

Let $S_{t,i} \in \{0,1\}$ be a binary random variable representing whether image $I_i$ satisfies the $t$-th feedback constraint.  For example, if the user's $t$-th comparison on attribute $m$ yields response $r = \text{``more''}$, then $S_{t,i} = 1$ if the database image $I_i$ has attribute $m$ more than the corresponding reference image $I_{t_f}$.  
We assume that the probability of an image satisfying a given constraint is independent of it satisfying another given constraint.
The probability that database image $I_i$ is relevant is the probability that it satisfies all $T$ feedback comparisons in $\mathcal{F}$:
\begin{equation}
P(y_i = 1 | I_i, \mathcal{F}) = \prod_{t=1}^T P(S_{t,i} = 1 | I_i, \mathcal{F}_t).
\label{eqn:probrel}
\end{equation}
For numerical stability, we replace the product above with a sum of log probabilities:
\begin{equation}
\log P(y_i = 1 | I_i, \mathcal{F}) = \sum_{t=1}^T \log P(S_{t,i} = 1 | I_i, \mathcal{F}_t).
\label{eqn:probrel2}
\end{equation}

The probability that an individual constraint is satisfied given that the user's response was $r$ for reference $I_{t_f}$ is:
\begin{equation}
 P(S_{t,i} = 1 | I_i, \mathcal{F}_t) =
  \begin{cases}
   P(A_m(I_i) > A_m(I_{t_f})) & \text{if } r = \text{``more''}\\
   P(A_m(I_i) < A_m(I_{t_f}))       & \text{if } r = \text{``less''}\\
   P(A_m(I_i) = A_m(I_{t_f})) & \text{if } r = \text{``equally''},\nonumber
  \end{cases}
\end{equation}
where $A_m(I)$ denotes the true strength of attribute $m$ in image $I$.  Note that we do not observe these true attribute values directly; rather, what we observe are the system's \emph{predicted} attribute values $a_m(I_i)$, which are necessarily imperfect.  While the predicted attribute ranks are a function of the true latent attribute strengths $A_m(I_i)$, they need not agree exactly.  Therefore, we estimate the probabilities required above by mapping the attribute predictions $a_m(\cdot)$ to probabilistic outputs.  We adapt Platt's method~\citep{Platt99} to the paired classification problem implicit in the large-margin ranking objective from Eqn.~\eqref{eq:joachimeq}.  Specifically, this yields:
\begin{equation}
{\small P(A_m(I_i) > A_m(I_{t_f})) = \frac{1}{1 + \exp(\alpha_m (a_m(I_i) - a_m(I_{t_f})) + \beta_m)},}\nonumber
\end{equation}
\begin{equation}
{\small P(A_m(I_i) < A_m(I_{t_f})) = 1 - P(A_m(I_i) > A_m(I_{t_f})), \hspace{0.05in} \text{and}}\nonumber
\end{equation}
\begin{equation}
\vspace{-0.2in}
{\small P(A_m(I_i) = A_m(I_{t_f})) = \frac{1}{1 + \exp(\gamma_m |a_m(I_i) - a_m(I_{t_f})| + \delta_m)}.}
\end{equation}
The sigmoid parameters are learned using the sets $O_m$ and $E_m$ from above.  In particular, to learn $\alpha_m$ and $\beta_m$, we use pairs with ``more'' judgments from $O_m$ as positive paired-instances, and ``less'' judgments as negative instances. For $\gamma_m$ and $\delta_m$, we use ``equally'' pairs from $E_m$ as positive labels, and both ``more'' and ``less'' responses from $O_m$ as negative instances.  We normalize these values so the three probabilities (``more''/``less''/``equally'') sum to 1.  

The relevance function defined above takes the user's input at face value.  Namely, if the user does \emph{not} comment on an attribute within the image, we assume we have no information about that attribute.  In other work, we explore how this assumption can be relaxed to learn the \emph{implicit} cues a user reveals in his/her attribute feedback~\citep{devi-iccv2013}.  For example, if a user elects to tell the system that his target is \emph{less shiny} than some reference reference $X$, and the reference image set the user saw contained another image $Y$ that is less shiny than $X$, then the system could infer that the target is \emph{not} less shiny than $Y$---otherwise, he would have provided that tighter constraint~\citep{devi-iccv2013}.

We stress that the proposed form of relative attribute feedback refines the search in ways that a straightforward multi-attribute query (e.g., as developed by~\cite{Kumar08}, \cite{Siddiquie11}, and \cite{Scheirer12}) cannot.  That is, if a user were to simply state the attribute labels of interest (``show me \emph{black} shoes that are \emph{shiny} and \emph{high-heeled}"), one can easily retrieve the images whose attribute predictions meet those criteria.  However, since the user's description is in absolute terms, it cannot evolve based on the retrieved images.  In contrast, with access to relative attributes as a mode of communication, for every new set of reference images returned by the system, a WhittleSearch user can further refine his description.  In addition, when a user states that a reference image has the attribute ``equally" to his target, he reveals more precise information than traditional binary relevance feedback.  In the former, we learn about the reference image's quality in the context of an individual attribute; in the latter, one learns only the coarse information that the image seems good or bad, across all attributes.

\subsection{Hybrid Feedback Approach}
\label{sec:hybrid}

So far, we have considered relative attribute feedback in isolation and discussed its advantages over traditional binary relevance feedback.  However, binary relevance feedback and relative attribute feedback can have complementary strengths: when reference images are nearly on target (or completely wrong in all aspects), the user may be best served by providing a simple binary relevance label.  Meanwhile, when a reference image is lacking only in certain describable properties, he may be better served by the relative attribute feedback.  Thus, it is natural to combine the two modalities, allowing a mix of feedback types at any iteration.

In a binary relevance feedback model, the user identifies a set of relevant images $\mathcal{R}$ and a set of irrelevant images $\mathcal{\bar{R}}$ among the current reference set $\mathcal{T}_t$.  In this case, the relevance scoring function is a classifier (or some other statistical model), and the binary feedback essentially supplies additional positive and negative training examples to enhance that classifier.  That is, the scoring function at iteration $t+1$ is trained with the data that trained the model at iteration $t$ \emph{plus} the images in $\mathcal{R}$ labeled as positive instances and the images in $\mathcal{\bar{R}}$ labeled as negative instances.

We can augment the WhittleSearch system with binary feedback to define a learned hybrid scoring function.  The basic idea is to learn a ranking function that unifies both relative attribute and binary feedback.  Let $\mathcal{C}_k \subset \mathcal{D}$ denote the subset of database images satisfying $k$ of the relative attribute feedback constraints, for $k=0,\dots,F$.  We define a set of ordered image pairs
\begin{equation}
O_s = \{\{\mathcal{R} \times \mathcal{\bar{R}}\} \cup \{\mathcal{C}_F \times \mathcal{C}_{F-1}\} \cup \dots \cup \{\mathcal{C}_1 \times \mathcal{C}_0\}\},
\end{equation}
where $\times$ denotes the Cartesian product.  This set $O_s$ reflects all the desired ranking preferences---that relevant images be ranked higher than irrelevant ones, and that images satisfying more relative attribute preferences be ranked higher than those satisfying fewer.  
Note that the subscript $s$ in $O_s$ distinguishes the set from those indexed by $m$ above, which were used to train relative attribute ranking functions in Section \ref{sec:learn-attributes}.

Using training constraints $O_s$
we learn a function that predicts \emph{relative image relevance} for the current user with the large-margin objective in Eqn.~\ref{eq:joachimeq}.  The result is a parameter vector $\bm{w}_s$ that serves as the hybrid scoring function.  Since there are many more pairs 
in $O_s$ 
that come from relative attribute feedback than from binary relevance feedback, we set the penalty on the binary feedback pairs to be inversely proportional to the fraction of such pairs in the set $O_s$.

\subsection{Active WhittleSearch with Attribute Pivots}
\label{sec:active}

Thus far, we have assumed that the user will freely select the feedback statements he wishes to give the system from among the top-ranked images.  This is the first \emph{user-initiated} variant of WhittleSearch, and it is most suited when a user wishes to browse at the same time he refines his own mental model of the target.  However, as argued above, when a user has a precise target in mind, it can be more beneficial to leave the choice of reference images for feedback to the image search system.  Thus, we next present an \emph{active} variant of WhittleSearch.

In Active WhittleSearch, the interaction mode involves a series of multiple-choice questions that the human user needs to answer, of the type: ``Is the image you are looking for \emph{more}, \emph{less}, (or \emph{equally}) $A$ than image $I$?", where $A$ is a semantic attribute and $I$ is an exemplar from the database being searched.  Our goal is to generate the series of such questions that will most efficiently narrow down the relevant images in the database, so that the user finds his target in few iterations.  To this end, at each iteration we will actively select a comparison for the user to provide, that is, the $(A,I)$ pair that yields the expected maximal information gain.  Rather than exhaustively search all database images as potential exemplars, however, we consider only a small number of \emph{pivot} exemplars---the internal nodes of binary search trees constructed for each attribute.  The output of the system is the list of database images, sorted by their predicted relevance.

As above, Active WhittleSearch also relies on predicted attribute values (Section \ref{sec:learn-attributes}) and a manner of updating the system's notion of relevance after each feedback statement it receives from the user (Section \ref{sec:relevance}).  It also relies on binary search trees, whose construction we explain next (Section \ref{sec:search-trees}).  Then, we introduce our active selection approach to determine which comparison should be requested next (Section \ref{sec:selection}) using the probabilistic model of image relevance defined in Section \ref{sec:relevance} above.

\subsubsection{Attribute Binary Search Trees}
\label{sec:search-trees}

For each attribute $m=1,\dots,M$, we construct a binary search tree.  The tree recursively partitions all the database images into two balanced sets, where the key at a given node is the median relative attribute value occurring within the set of images passed to that node.  To build the $m$-th attribute tree, we start at the root with all database images, sort them by their attribute values $a_m(I_1),\dots,a_m(I_N)$, and identify the median value.  Let $I_p$ denote the ``pivot" image---the one that has the median attribute strength.  Those images exhibiting the attribute less than $I_p$, i.e., all $I_i$ such that $a_m(I_i) \leq a_m(I_p)$, are passed to the left child, while those exhibiting the attribute more, i.e., $a_m(I_i) > a_m(I_p)$, are passed to the right child.  Then the splitting repeats recursively, each time storing the next pivot image and its relative attribute value at the appropriate node.

Note that both the relative attribute ranker training and the search tree construction are offline procedures; they are performed once, before handling any user queries.

Already, one could imagine a search procedure that walks a user through one such attribute tree, at each successively deeper level requesting a comparison to the pivot, and then eliminating the appropriate portion of the database depending on whether the user says ``more" or ``less".  However, there are two problems with such a simple approach.  First, we cannot assume that the attribute predictions are identical to the attribute strengths a user will perceive; thus, a hard pruning of a full sub-tree is error-prone.  Second, it fails to account for the variable information gain that could be achieved depending on which attribute is explored at any given round of feedback.  Therefore, we use the probabilistic representation of whether images satisfy the comparison constraints, as defined in Section \ref{sec:relevance}, and we use the pivots to limit the pool of candidate images that are evaluated for their expected information gain, as we will explain next.

\subsubsection{Actively Selecting an Informative Comparison}
\label{sec:selection}

Our system maintains a set of $M$ current pivot images (one per attribute tree) at each iteration, which we denote by \linebreak $\mathcal{P}=\{I_{p_1},\dots,I_{p_M}\}$, where $\mathcal{P} \subset \mathcal{D}$.  The pivots are initially the root pivot images from each tree.  During active selection, our goal is to identify the pivot in this set that, once compared by the user to his target, will most reduce the entropy of the relevance predictions on all database images in $\mathcal{D}$.  Note that selecting a pivot corresponds to selecting both an image as well as an attribute along which we want it to be compared.  That is, $I_{p_m}$ refers to the pivot for attribute $m$.  

\paragraph{Entropy reduction objective.}

Given the feedback history $\mathcal{F}$, we want to predict the information gain across all $N$ database images for each pivot in $\mathcal{P}$.  We will request a comparison for the pivot that most reduces the total relevance entropy over all images---or equivalently, the pivot that minimizes the expected entropy when used to augment the current set of feedback constraints.

The entropy based on the feedback thus far is:
\begin{equation}
H(\mathcal{F}) = - \sum_{i=1}^N \sum_{\ell} P(y_i=\ell | I_i, \mathcal{F}) \log P(y_i = \ell | I_i, \mathcal{F}),
\end{equation}
where $\ell \in \{0,1\}$.  Let $R$ be a random variable denoting the user's response, $R \in \{\text{``more"},\text{``less"}, \text{``equally"}\}$.  We select the next pivot for comparison as:
\begin{equation}
I_p^\ast = \argmin_{I_{p_m} \in \mathcal{P}} \sum_{r} P(R = r | I_{p_m}, \mathcal{F})~~H(\mathcal{F} \cup (I_{p_m}, r)),\label{eqn:bestpivot}
\end{equation}
where $H(\mathcal{F} \cup (I_{p_m}, r))$ denotes the entropy computed on the accumulated feedback when it is further augmented with the hypothetical response $r$ on pivot image $I_{p_m}$, and $P(R = r | I_{p_m}, \mathcal{F})$ is the likelihood of the user giving the response $r$.  In other words, the most informative pivot---the one the user should next compare his target image to---is the pivot that most reduces the expected entropy.

\paragraph{User response likelihood.}
\label{sec:user_resp}

Optimizing Eqn.~\ref{eqn:bestpivot} requires estimating the likelihood of each of the three possible user responses to a question we have not issued yet.  We develop three possible strategies to estimate it.  In each case, we use cues from the available feedback history to form a ``proxy" for the user, essentially borrowing the probability that a new constraint is satisfied from previously seen feedback.

For the first strategy, which we call \textsc{All Relevant}, we use all relevant database images as the proxy.  The assumption is that the images that are relevant to the user thus far are (on the whole) more likely to satisfy the user's next feedback than those that are irrelevant.  This is reminiscent of standard practice in active classifier training, where posteriors estimated with the current classifier are used as weights in the expected entropy reduction of acquiring a new label.  Ideally we would average the $P(S_{c,i} = 1 | I_i, \mathcal{F}_c)$ values among only the relevant images $I_i$, where $c$ indexes the candidate new feedback for a (yet unknown) user response $R$.  Of course, we can only \emph{predict} relevance, so we compute the weighted probability of each possible response $R$:
\begin{equation}
{\small P_{all}(R=r | I_{p_m},\mathcal{F}) = \frac{1}{N} \sum_{i=1}^N P(y_i=1 | I_i, \mathcal{F}) P(S_{c,i} = 1 | I_i, \mathcal{F}_c),}
\end{equation}
where the \emph{all} subscript stands for \textsc{All Relevant}.

The second strategy, which we call \textsc{Most Relevant}, is similar, but uses only our current best guess for the target image as the proxy:
\begin{equation}
P_{most}(R=r | I_{p_m},\mathcal{F}) = P(S_{c,b} = 1 | I_b, \mathcal{F}_c),
\end{equation}
where $I_b$ is the database image that maximizes $P(y_i=1 | I_i, \mathcal{F})$, for $i=1,\dots,N$.

The third strategy, which we call \textsc{Similar Question}, \linebreak examines all previously answered feedback requests, and copies the answer from the question that is most similar to the new one.  We define question similarity in terms of the Euclidean distance between the pivot images' descriptors plus the similarity of the two attributes involved in either question.  We quantify the latter by the Kendall's $\tau$ correlation between the ranks they assign to a set of validation images. For example, this reflects that \emph{feminine} and \emph{heel height} are more aligned than \emph{feminine} and \emph{grayness}.  Let $r_k^\ast$ denote the response to the most similar question $k$ found in the history $\mathcal{F}$ for the new pivot $I_{p_m}$ under consideration.  Then we have:
\begin{equation}
P_{question}(R=r | I_{p_m},\mathcal{F}) =
\begin{cases}
   1  & \text{if } r = r_k^\ast\\
   0 & \text{otherwise }.
  \end{cases}
\end{equation}
We evaluate all three likelihood strategies in the results.

\begin{figure}[t]
\centering
\includegraphics[width=0.9\linewidth]{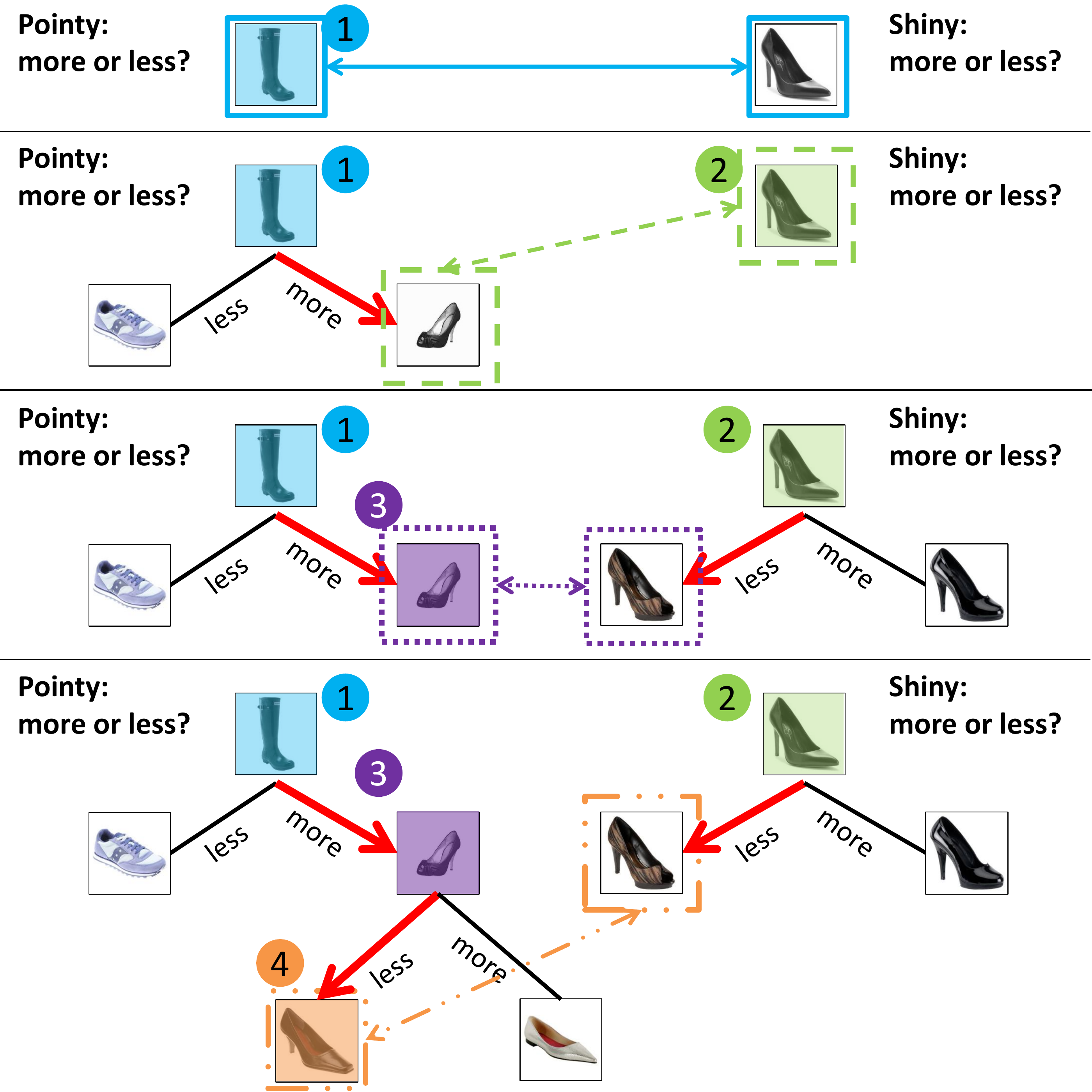}
\caption{The Active WhittleSearch variant requests feedback on images that elicit the most information, using binary search trees to focus the active selection.  In this sketch, $M=2$ attribute trees are shown.  Images with the same color outline are the pairs considered at each round, and the number in this color marks the image chosen at this round.  Red arrows denote the user's responses.  Here, first the user is asked to compare his target to the boot pivot (1) in terms of pointiness; then he is asked to compare it to (2) in terms of shininess, followed by (3) in terms of pointiness, and so on.  Best viewed in color.}
\label{fig:approach-AWS}
\end{figure}

\paragraph{Recap of Active WhittleSearch interaction loop.}

At each iteration, we present the user with the pivot selected with Eqn.~\ref{eqn:bestpivot} and request the specified attribute comparison.  Then, we (1) use his response to update $\mathcal{F}$ with that additional image-attribute-response constraint, and (2) either replace the pivot in $\mathcal{P}$ for that attribute with its appropriate child pivot (i.e., the left or right child in the binary search tree if the response is ``less" or ``more", respectively) or terminate the exploration of this tree (if the response is ``equally'').  Note that this means that the set of pivots consists of pointers into the binary trees at \emph{varying} levels.  See Figure~\ref{fig:approach-AWS}.  This is because our active selection criterion considers which attribute will most benefit from more refined feedback at any point in time.  In contrast, a simpler solution that alternates between the attribute trees in sequence need not reduce uncertainty as efficiently, as we will show in the results.

Finally, the approach iterates until the user is satisfied with the top-ranked results, or until all of the attribute trees have bottomed out to an ``equally" response from the user (in which case, our method can gain no further knowledge about the target given the available attribute vocabulary).

The cost of our selection method per round of feedback is $O(MN)$, where $M$ is the size of the attribute vocabulary, $N$ is the database size, and $M \ll N$.  
For each of $O(M)$ pivots which can be used to complement the feedback set, we need to evaluate expected entropy for all $N$ images.
In contrast, a traditional information gain approach would scan all database items paired with all attributes, requiring $O(MN^2)$ time.
The proposed binary search trees exploit the ordinal values of relative attributes to make this complexity reduction possible.

\subsection{Discussion}
\label{sec:discussion-WS-vs-AWS}

Having described both WhittleSearch variants, we can now compare and contrast them in detail.  Recall that feedback is user-initiated in the first variant: the system presents the user with the top-ranked current results, and the user freely chooses those on which he wishes to provide comparative feedback.  The second variant, Active WhittleSearch, is \linebreak system-initiated: the system asks the user for a visual comparison between the envisioned target image and an actively selected reference image along a specific attribute.  Both variants have potential advantages that are revealed under different scenarios.  

Active WhittleSearch makes a choice that is optimal with respect to the knowledge that the image search system possesses. This can be likened to a situation where we rely on a student's own understanding of what he knows in order to improve his knowledge. However, unlike WhittleSearch, the set of images that is shown to the user for feedback is often disjoint from those that are ranked highest by the system. Therefore, the user must separately examine the images for feedback and the image results.

In contrast, WhittleSearch gives the human user several options about the reference images and attributes on which to comment. Therefore, the performance of the system depends both on the choices that the user makes, as well as the correctness of the response that the user gives on the chosen pairing of image and attribute. In this case, we rely on the human ``teacher'' to know what additional information to give to the system ``learner''.  WhittleSearch also requires more time for the completion of one feedback statement compared to Active WhittleSearch, since it requires the user to examine a set of options and choose among them.

In cases when the user does not wish to spend much time considering which image and attribute to comment on, we expect that Active WhittleSearch will be preferred. For example, the user might choose to comment on those comparisons which are most obvious, which might not be very informative to the system. However, if the user is careful and experienced enough with the system to pick informative comparisons, WhittleSearch can perform better.  For example, the user might see a unique attribute which is important for discriminating between relevant and irrelevant images, which the system has not asked about yet. This will be particularly important if there is a large discrepancy between the human perception of an attribute and the system ranking for this attribute, in which case the entropy reduction estimates might be inaccurate.
 
Another factor which affects how well the two versions of WhittleSearch perform is the number of feedback statements that the system has received so far. As we will show in our results (Section \ref{sec:results-AWS}), the entropy-based selection criterion is most crucial early on in the iterative cycle.  Thus, we expect the advantage of Active WhittleSearch over WhittleSearch to be stronger in the first few iterations.

Finally, the level of specificity of the user's target might affect WhittleSearch and Active WhittleSearch's comparative performance as well. If the user is simply browsing, WhittleSearch might be preferable as it gives him more freedom to explore the current results and refine or terminate the search, depending on the precise qualities of the desired target. 
For example, a user shopping for a product with only a vague preconception of what is desired may be best suited by WhittleSearch.
However, if the user has a very specific target in mind, Active WhittleSearch might be more helpful, as the use of binary search trees helps narrow down the search to the exact range of the attribute value distribution that matches the ``signature'' of the target image. The feasibility of browsing can be affected by the size of the search interface.  For example, it might be harder to browse reference images on a small mobile phone screen, which speaks in favor of eliminating user choice for the feedback statements, and trying to pinpoint the exact object that the user has in mind.

\begin{figure}[h]
\includegraphics[width=1\linewidth]{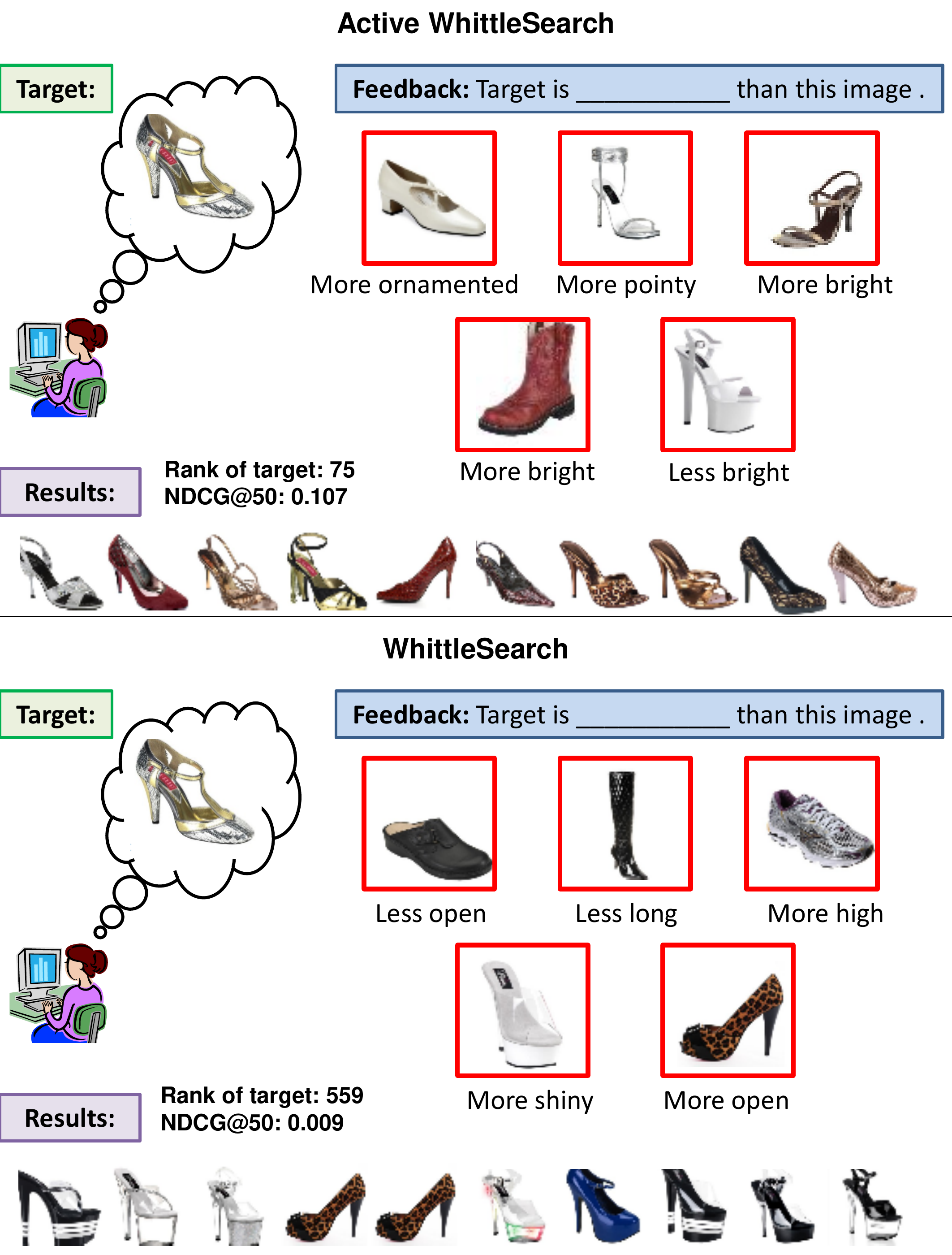}
\caption{An example where Active WhittleSearch outperforms WhittleSearch.  Observe how the active selection focuses on the two most distinctive features of this shoe, namely its color and ornaments, which the human user fails to do.  The user gives feedback that is obvious yet not very discriminative; most shoes are \emph{less long-on-the-leg} than a boot, and many shoes in this dataset are \emph{higher at the heel} than a running shoe.  See Section \ref{sec:results} for all implementation details leading to this result.}
\label{fig:qual_AWS_4}
\end{figure}

\begin{figure}[h]
\includegraphics[width=1\linewidth]{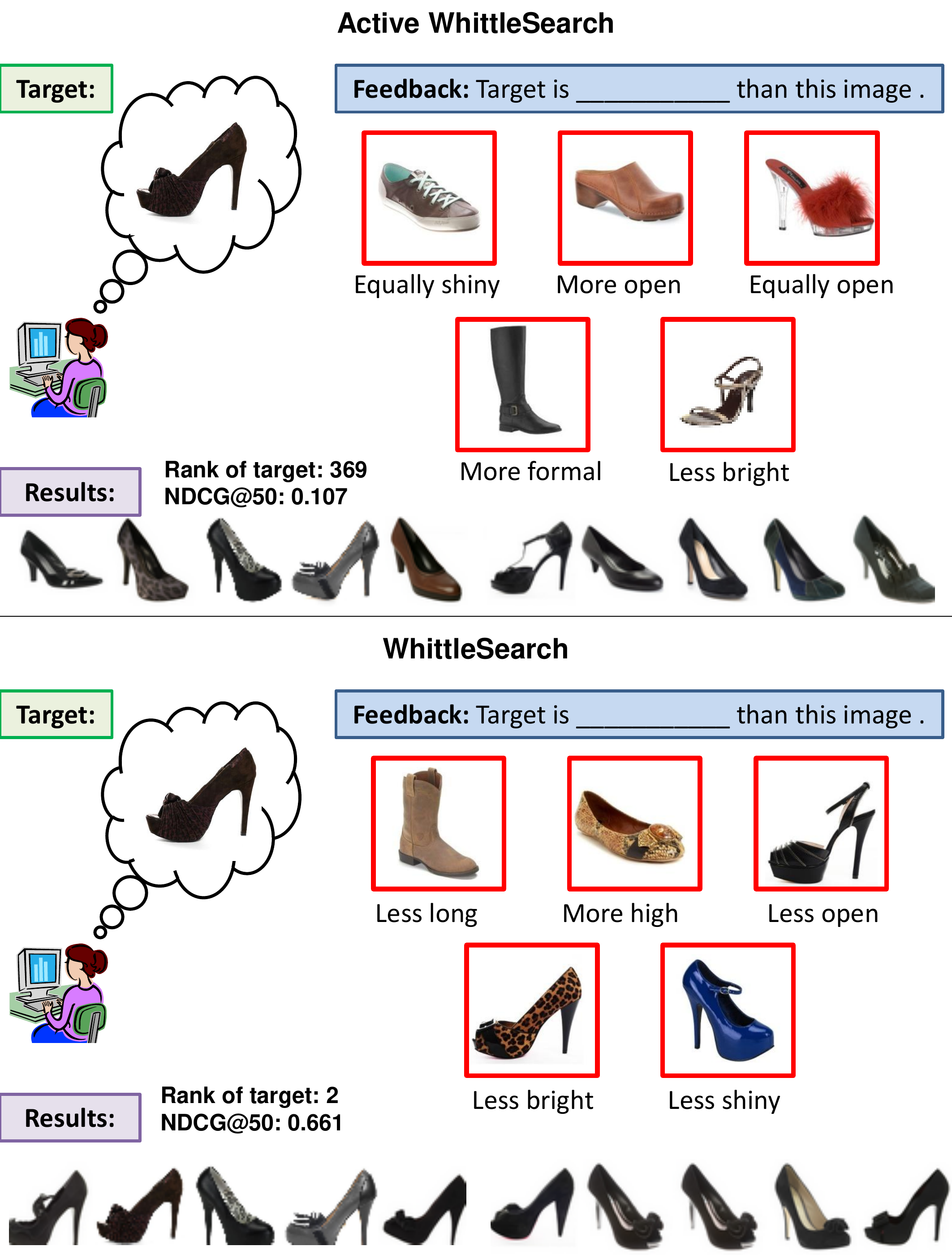}
\caption{An example where WhittleSearch outperforms Active WhittleSearch.  While Active WhittleSearch does a fair job, this particular user of WhittleSearch gave very useful feedback, which allowed the system to rank the target image nearly at the top of the results page.  See Section \ref{sec:results} for all implementation details leading to this result.}
\label{fig:qual_AWS_5}
\end{figure}

Figures \ref{fig:qual_AWS_4} and \ref{fig:qual_AWS_5} show two qualitative comparisons of the two WhittleSearch variants, which illustrate some of the tradeoffs discussed.  
The first figure shows user-chosen feedback that does not point out the most distinctive features of the target image, while the second shows particularly valuable user-chosen feedback.

\section{Experimental Results}
\label{sec:results}

We first explain our experimental setup in Section \ref{sec:design}.  
In Section~\ref{sec:results_WS} we analyze how the proposed relative attribute feedback can enhance image search compared to classic binary feedback, and study which factors influence their behavior.  Then, in Section~\ref{sec:results-AWS} we compare our active selection method in the Active WhittleSearch variant to alternative selection strategies to demonstrate its benefits.
Finally, in Section \ref{sec:results-comparison}, we experimentally compare WhittleSearch and Active WhittleSearch.

\subsection{Experimental Design}
\label{sec:design}

\paragraph{Datasets.}

We use three datasets in order to validate our approach in diverse domains of interest: finding products, people, and scenes.  The datasets are:
\begin{itemize}
\item The \textbf{Shoes} dataset from the Attribute Discovery Dataset \citep{Berg10}, which contains 14,658 shoe images belonging to 10 shoe categories collected from the website \texttt{like.com}.  We augment the data with 10 relative attributes---\emph{pointy at the front}, \emph{open}, \emph{bright in color}, \emph{covered with ornaments}, \emph{shiny}, \emph{high at the heel}, \emph{long on the leg}, \emph{formal}, \emph{sporty}, and \emph{feminine}.
\item The Public Figures dataset of human faces \citep{Kumar09} (\textbf{Faces}). We use the subset from \citep{Parikh11b}, which contains 772 images from 8 people and 11 attributes---\emph{masculine-looking}, \emph{white}, \emph{young}, \emph{smiling}, \emph{chubby}, \emph{visible forehead}, \emph{bushy eyebrows}, \emph{narrow eyes}, \emph{pointy nose}, \emph{big lips}, and \emph{round face}.
\item The Outdoor Scene Recognition dataset of natural scenes \citep{Oliva01} (\textbf{Scenes}), which consists of 2,688 images from 8 categories and 6 attributes---\emph{natural}, \emph{open}, \emph{perspective}, \emph{large objects}, \emph{diagonal plane}, and \emph{close depth} \citep{Parikh11b}.
\end{itemize}

\paragraph{Features.}

For image features $\bm{x}$, we use GIST~\citep{Oliva01} and LAB color histograms for Shoes and Faces, and GIST alone for Scenes.  We omit color for Scenes because we expect that the majority of scene attributes cannot be captured with color features.  The GIST descriptor captures the overall texture of the image, summarizing gradient orientations in a grid of spatially localized cells.  The color histogram summarizes the color distribution in the image, offering complementary information to the GIST descriptor.
For Shoes, we concatenate a 960-dimensional GIST feature vector (4 blocks and 8-8-4 orientations per scale) and a 30-dimensional color feature vector (10 bins). For Scenes, we use a 512-dimensional GIST vector. For Faces, we concatenate a 512-dimensional GIST vector and a 30-dimensional color vector.

\paragraph{Methodology.}

For each query we select a random \emph{target image} and score how well the search results match that target after feedback.  This target stands in for a user's mental model; it allows us to prompt multiple subjects for feedback on a well-defined visual concept, and to precisely judge how accurate results are. This part of our methodology is key to ensure consistent data collection and formal evaluation.

We use two evaluation metrics: (1) the ultimate \emph{percentile rank} assigned to the user's target image, which measures the fraction of database images ranked \emph{below} the true target, and (2) the \emph{correlation} between the full ranking computed by the method's relevance scoring function and a ground truth ranking that reflects the perceived relevance of all images in $\mathcal{D}$.  For both metrics, higher scores are better.

The correlation metric captures not only where the target itself ranks, but also how similar to the target the other top-ranked images are.  We form the ground truth relevance ranking by sorting all images in $\mathcal{D}$ by their distance to the given target.  To ensure this distance reflects \emph{perceived} relevance, we learn a metric based on human judgments.  Specifically, we show 750 triplets of images  $(i, j, k)$ from each dataset to seven Mechanical Turk human subjects, and ask whether images $i$ and $j$ are more similar, or images $i$ and $k$.   Using their responses, we learn a linear combination of the image and attribute feature spaces that respects these constraints via~\citep{Joachims02}.  Our ground truth rankings thus mimic human perception of image similarity.  To score correlation, we use Normalized Discounted Cumulative Gain at top $K$ (NDCG@K)~\citep{Kekalainen02}.  This is a standard information retrieval metric that scores how well the predicted ranking and the ground truth ranking agree, while emphasizing items ranked higher.  We use $K=50$, based on the number of images visible on a page of image search results.

\begin{figure*}[t]
\centering
\begin{tabular}{ccc}
\includegraphics[width=0.25\linewidth]{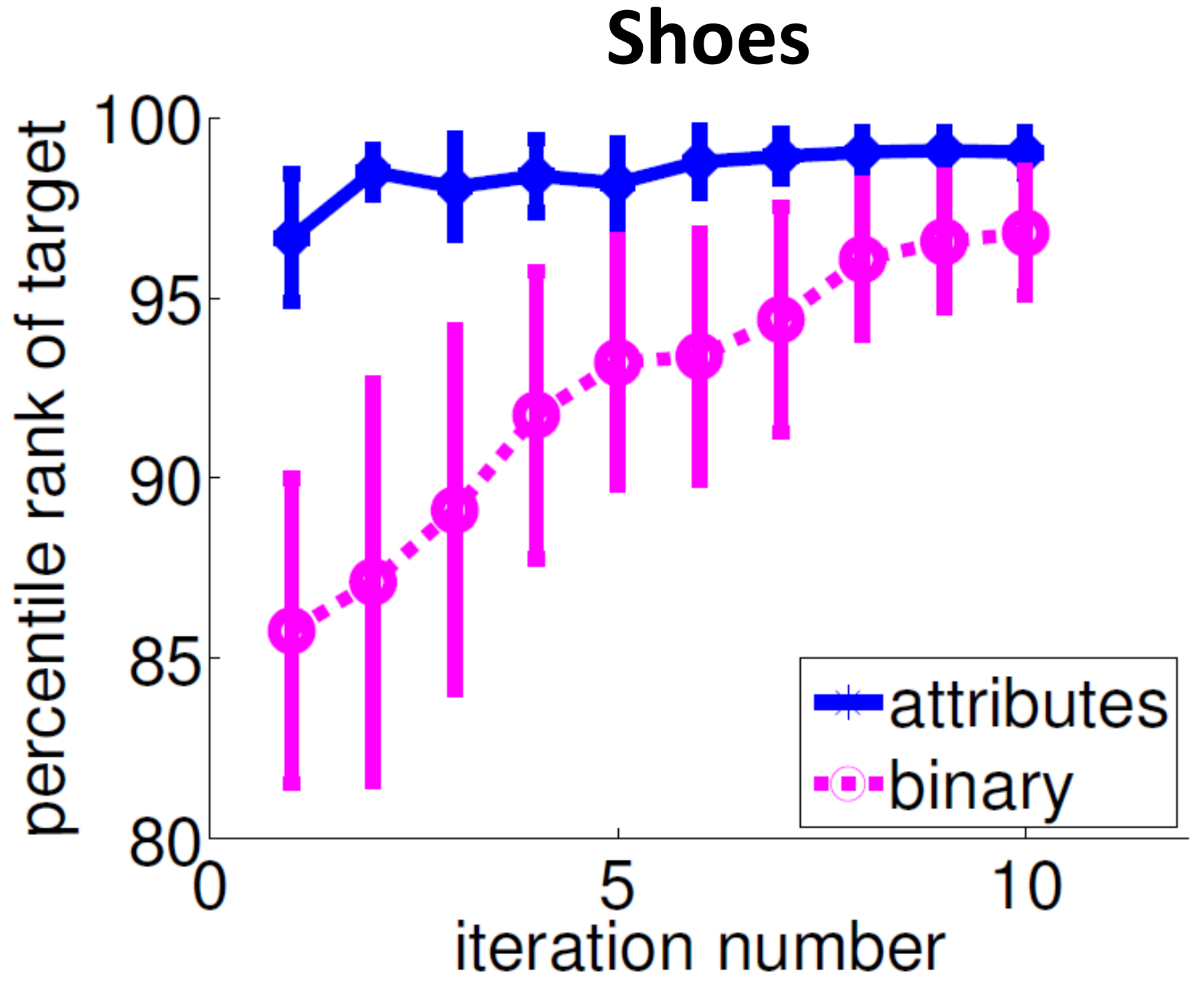}&
\includegraphics[width=0.25\linewidth]{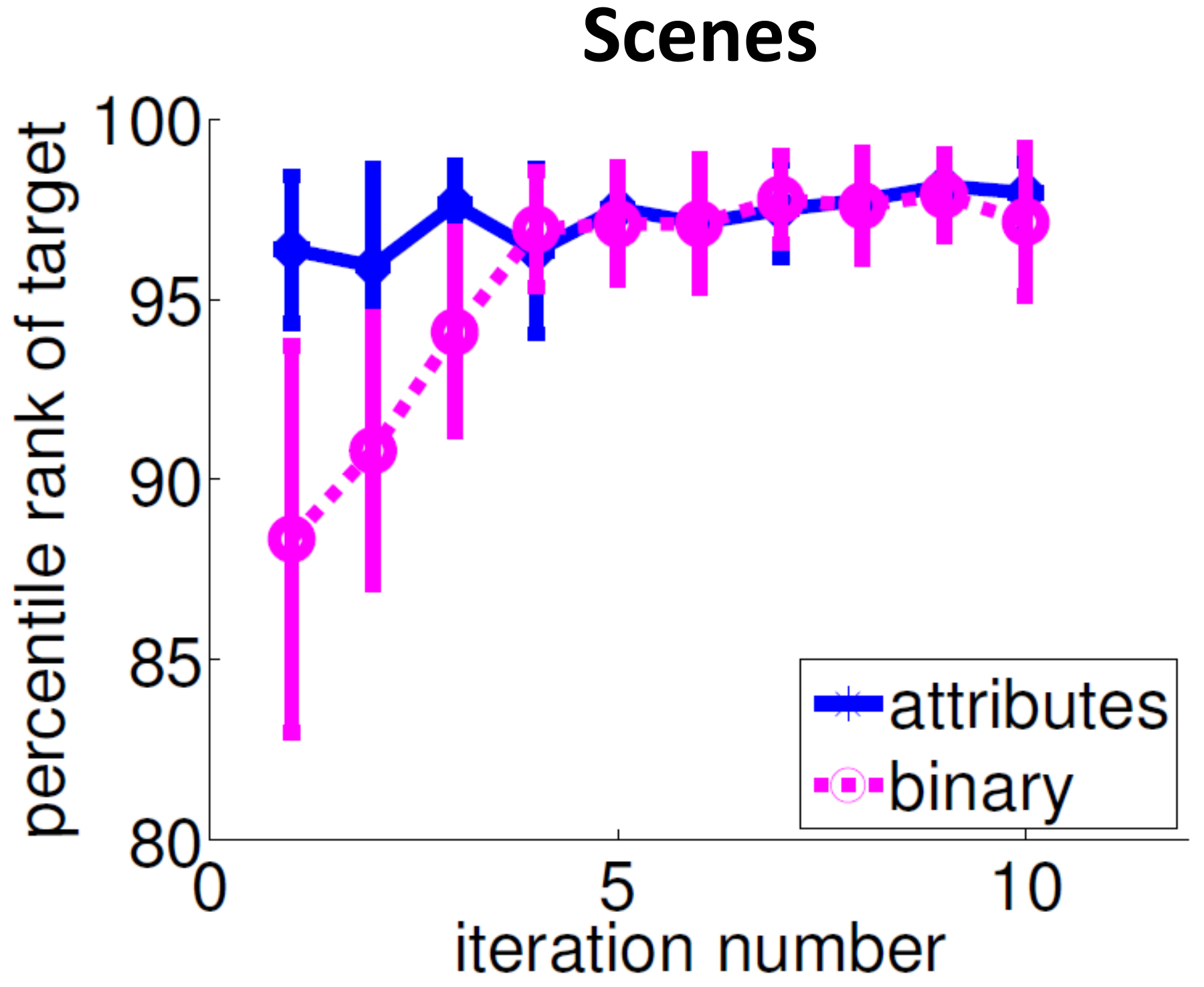}&
\includegraphics[width=0.25\linewidth]{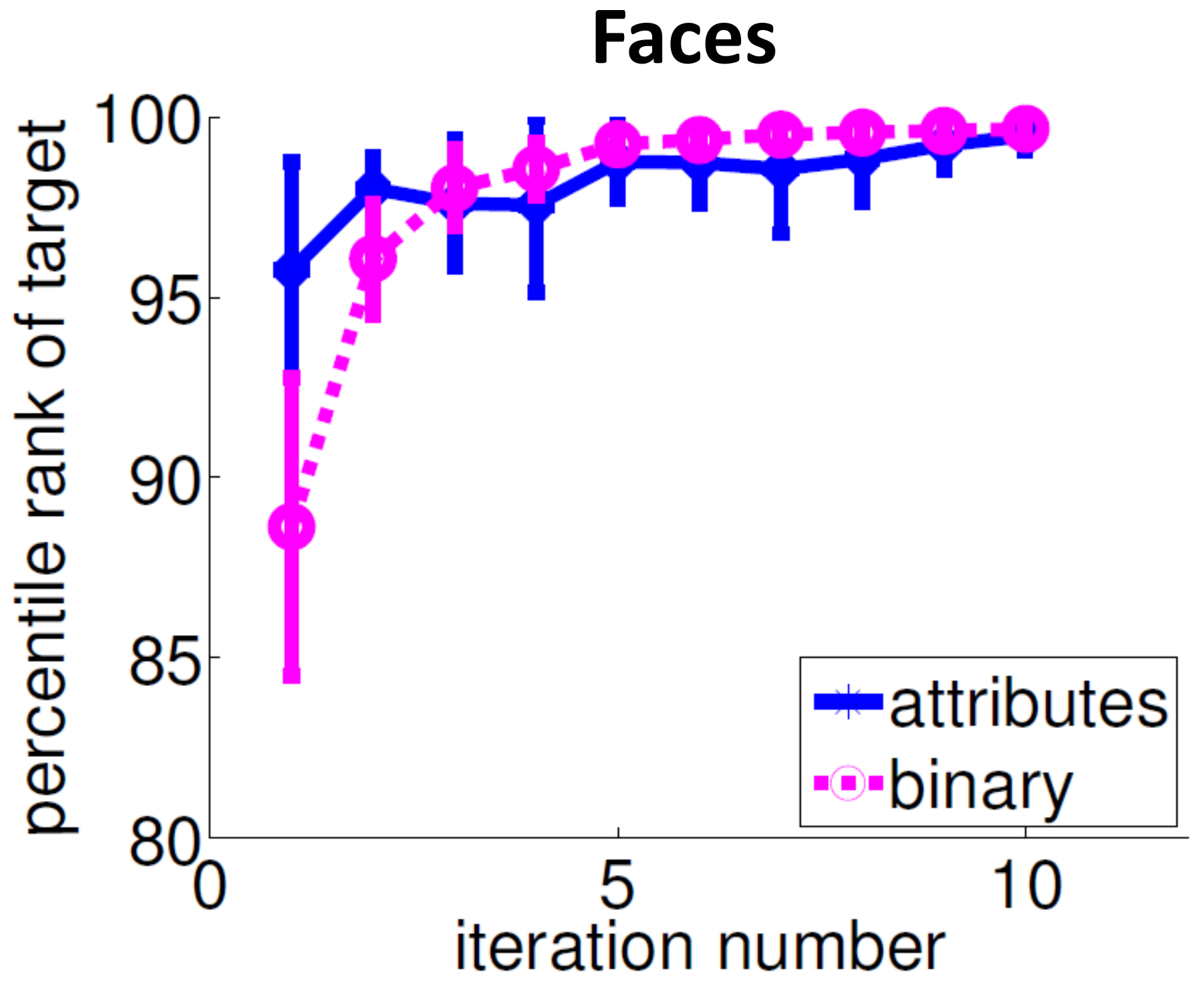}
\end{tabular}
\caption{Iterative search with WhittleSearch vs. traditional binary relevance feedback on three datasets.  We show accuracy (percentile rank of the target image) as a function of the number of iterations of feedback.   Our method often converges on the target image more rapidly.}
\label{fig:323}
\end{figure*}

\paragraph{Baseline.} 

The key baseline against which we compare WhittleSearch is traditional binary relevance feedback.  This baseline is intended to represent existing approaches such as \citep{Cox00,Ferecatu07,Rui98,Tieu00}.  While a variety of classifiers have been explored in such previous systems, we employ a support vector machine (SVM) classifier for the binary feedback model due to its strong performance in practice.  Thus, the relevance scoring function for the binary feedback baseline is the magnitude of the SVM output.  (We defer the definition of the additional baselines against which we test Active WhittleSearch until Section~\ref{sec:results-AWS}.)

\begin{figure*}[t]
\centering
\begin{tabular}{ccc}
\includegraphics[width=0.25\linewidth]{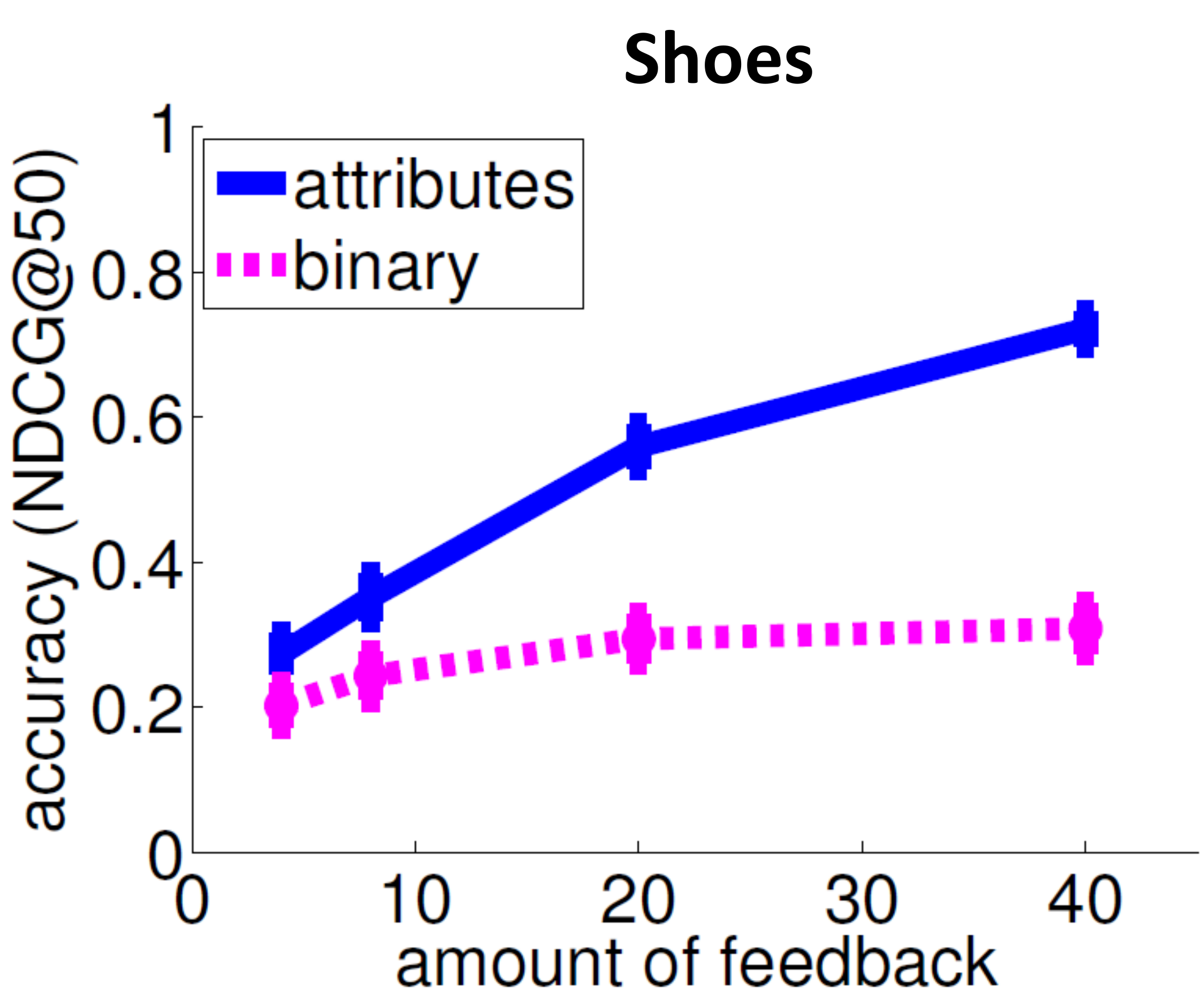}&
\includegraphics[width=0.25\linewidth]{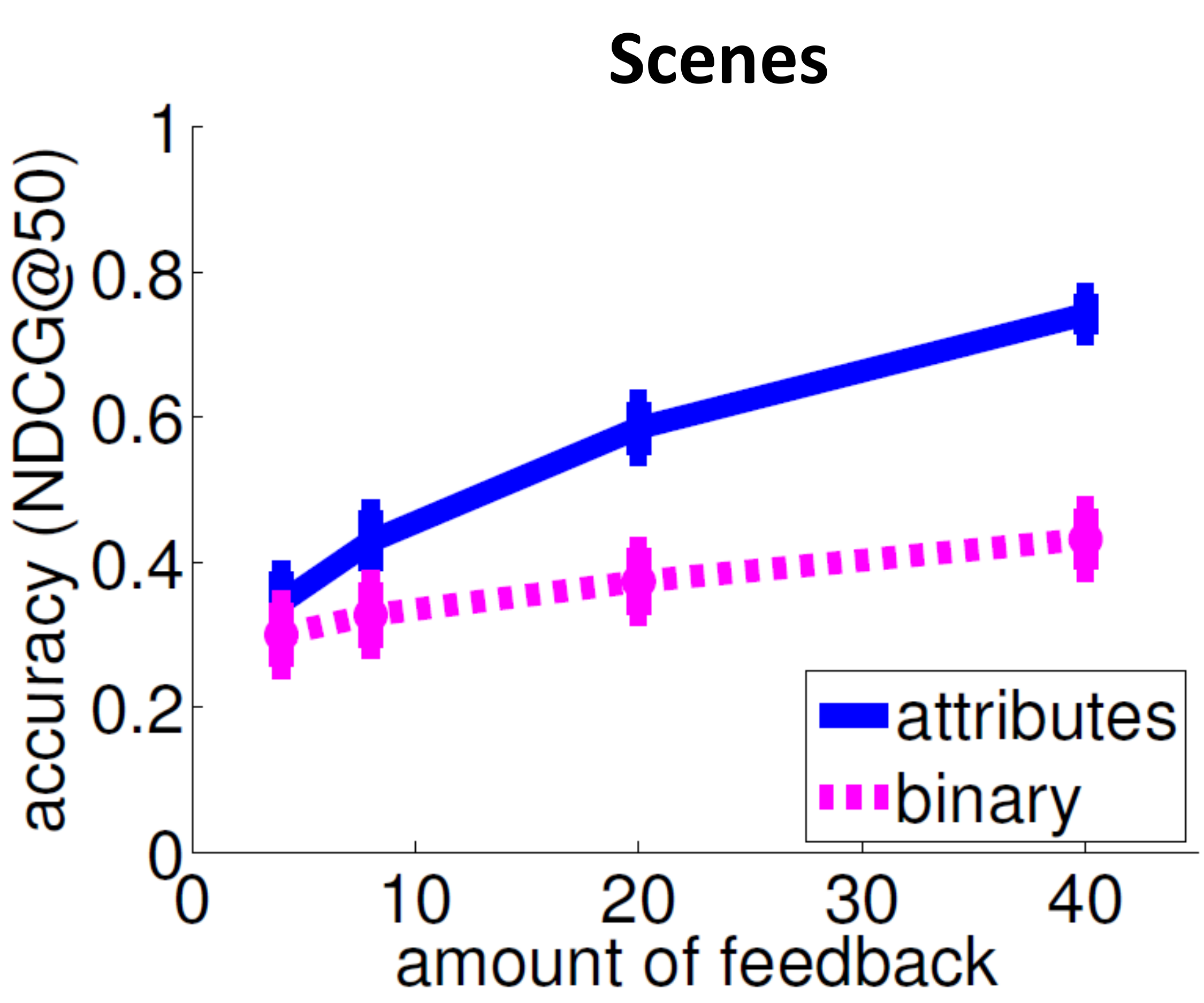}&
\includegraphics[width=0.25\linewidth]{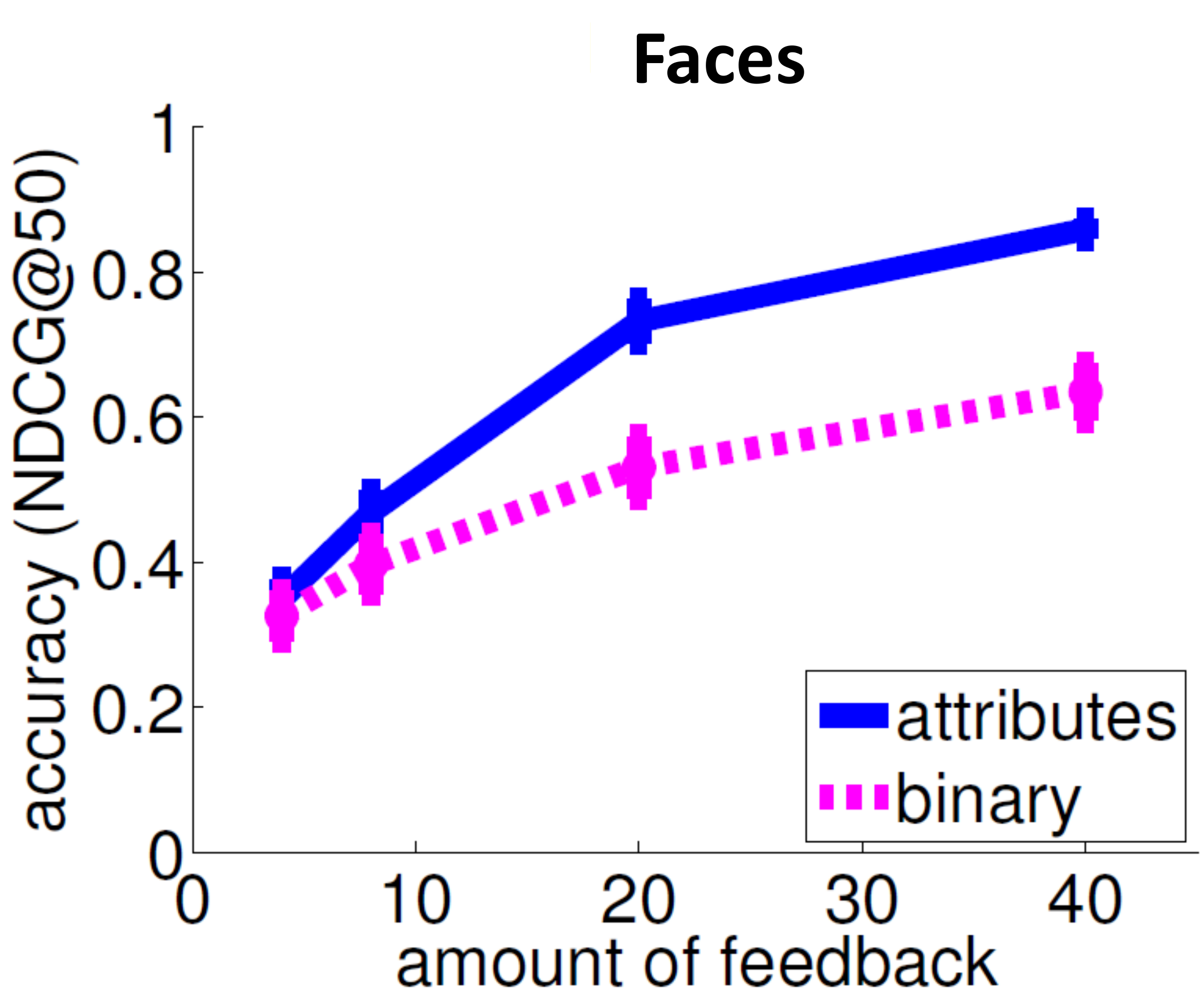}
\end{tabular}
\caption{Ranking accuracy as a function of amount of feedback.  While more feedback enhances both our method and the traditional binary relevance feedback approach, the proposed attribute feedback yields faster gains per unit of feedback.}
\label{fig:705}
\end{figure*}

\subsection{WhittleSearch Results}
\label{sec:results_WS}

% ----- feedback generation part -----
We use Mechanical Turk to gather human feedback for our relative attribute method and the binary feedback baseline.  We pair each target image with 16 reference images.  For our method we ask, ``Is the target image more or less $\langle$attribute name$\rangle$ than the reference image?'' (for each $\langle$attribute name$\rangle$), while for the baseline we ask, ``Is the target image similar to or dissimilar from the reference image?''  We also request a confidence level for each answer, as shown above in Figure~\ref{fig:relative-supervision}. We get each pair labeled by up to five workers and use majority voting to reduce noise.  When sampling from these constraints to impose feedback, we take those that have the highest average confidence levels, assuming that a user will select that response of which he is most confident.

Since the human annotations are costly, for certain studies below we generate feedback automatically.  For relative constraints, we randomly sample constraints based on the predicted relative attribute values, checking how the target image relates to the reference images.  In other words, the simulated user randomly chooses an attribute and one of the $n$ top-ranked images at that round, and compares his target image to the chosen reference image along the given attribute dimension.  For example, if the target's predicted ``shininess'' is 0.5 and the reference image's ``shininess'' is 0.6, then a valid constraint is that the target is ``less shiny'' than that reference image.  For binary feedback, we analogously sample positive/negative reference examples based on their image feature distance to the true target.  In particular, we sort the $n$ currently top-ranked in terms of their Euclidean distance in raw feature space to the target image.  We then generate constraints that say the top quartile of these images are ``similar to'' the target image, while the bottom quartile are ``dissimilar from'' the target.

When scoring rank, we add Gaussian noise to the predicted attributes (for our method) and the SVM outputs (for the baseline), to coarsely mimic human uncertainty in constraint generation.  The automatically generated feedback is a good proxy for human feedback since the relative predictions are explicitly trained to represent human judgments. It allows us to test performance on a larger scale.

% ----- end of feedback generation -----

First we evaluate the core WhittleSearch system with user-initiated feedback.  These results aim to establish the value of relative attribute feedback compared to traditional binary relevance feedback.  Since there is no active selection and we do not need to estimate entropy reduction in these results, we simplify the probabilistic relevance function in Eqn.~\ref{eqn:probrel2} to use binary values for the probabilities $P(S_{t,i} = 1 | I_i, \mathcal{F}_t)$, such that the relevance function simply counts the number of constraints satisfied by a database image $I_i$.  Specifically, this corresponds to defining:
\begin{equation}
P(A_m(I_i) > A_m(I_{t_f})) = [a_m(I_i) > a_m(I_{t_f})], \text{and}\nonumber
\end{equation}
\begin{equation}
P(A_m(I_i) < A_m(I_{t_f})) = [a_m(I_i) < a_m(I_{t_f})],
\end{equation}
where the brackets denote Iverson bracket notation.

\paragraph{Impact of iterative feedback.}

First we examine how the rank of the target image improves as the methods iterate.  Both methods start with the same random set of 16 reference images, and then iteratively obtain eight automatically generated feedback constraints, each time re-scoring the data to revise the top reference images.  To ensure new feedback accumulates per iteration, we do not allow either method to reuse a reference image.

Figure~\ref{fig:323} shows the results, for 50 such queries.  Our \linebreak method outperforms the binary feedback baseline for all \linebreak datasets, more rapidly converging on a top rank for the target image.  On Faces our advantage is slight, however.  We suspect this is due to the strong category-based nature of the Faces data, which makes it more amenable to binary feedback; adding positive labels on exemplars of the same person as the target image is quite effective.  In contrast, on Scenes and Shoes, where images have more fluid category boundaries, our advantage is much stronger.  The searches tend to stabilize after 2-10 rounds of feedback.  The run-times for our method and the baseline are similar.

\paragraph{Impact of amount of feedback.}

Next we analyze the impact of the amount of feedback, using automatically generated constraints.  Figure~\ref{fig:705} shows the rank correlation results for 100 queries.  These curves show the quality of all top-ranked results as a function of the amount of feedback given in a single iteration.  Recall that a round of feedback consists of a relative attribute constraint or a binary label on one image, for our method or the baseline, respectively.  For all datasets, both methods clearly improve with more feedback.  However, the precision enabled by our attribute feedback yields a greater ``bang for the buck''---higher accuracy for fewer feedback constraints.  The result is intuitive, since with our method users can better express \emph{what about} the reference image is (ir)relevant to them, whereas with binary feedback they cannot.

A multi-attribute query baseline that ranks images by how many binary attributes they share with the target image achieves NDCG scores 40\% weaker on average than our method when using 40 feedback constraints.  This result supports our claim that binary attribute search lacks the expressiveness of iterative relative attribute feedback.

\begin{figure*}[t]
\centering
\begin{tabular}{cccc}
\includegraphics[width=0.22\textwidth]{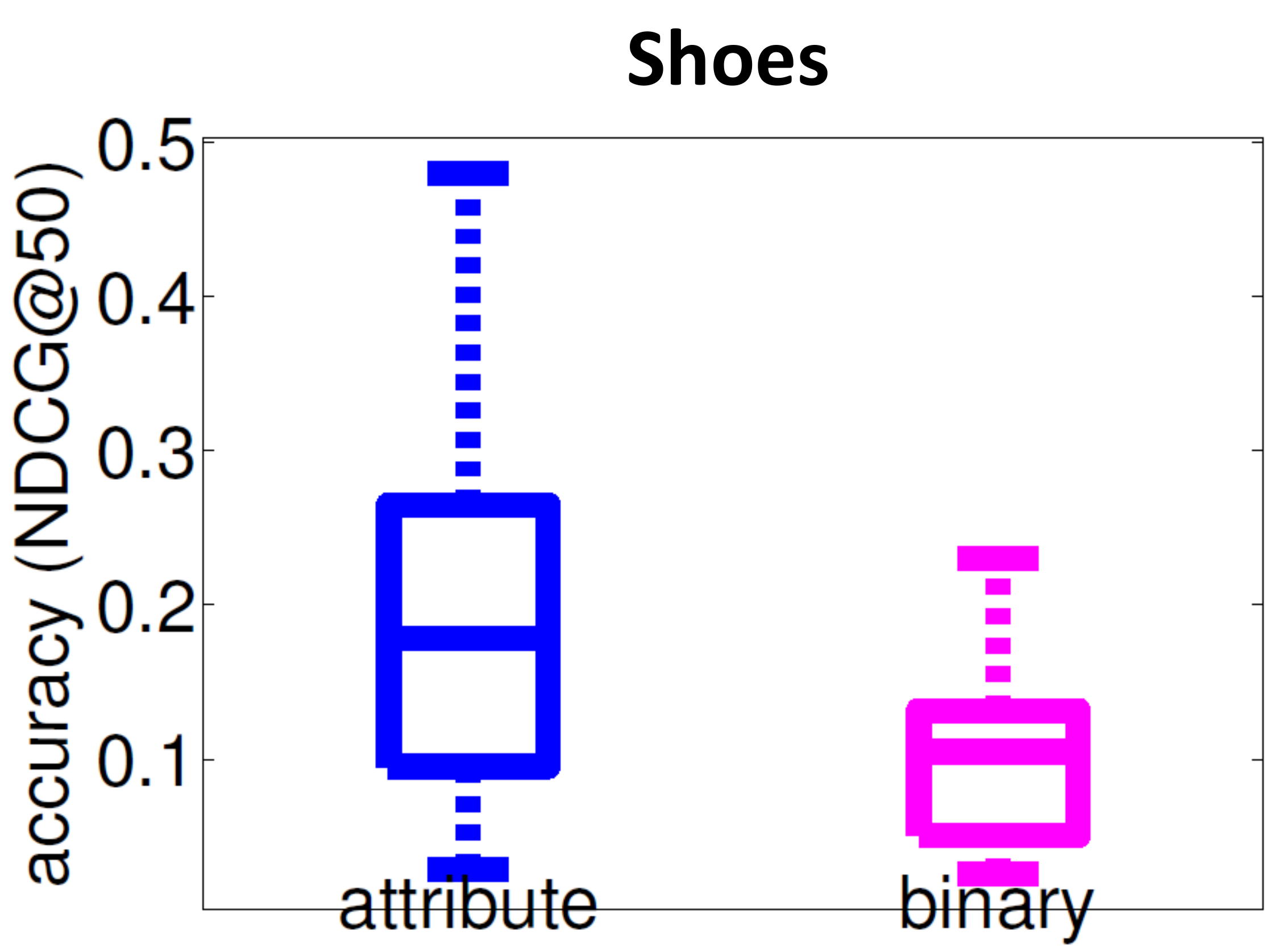}&
\includegraphics[width=0.22\textwidth]{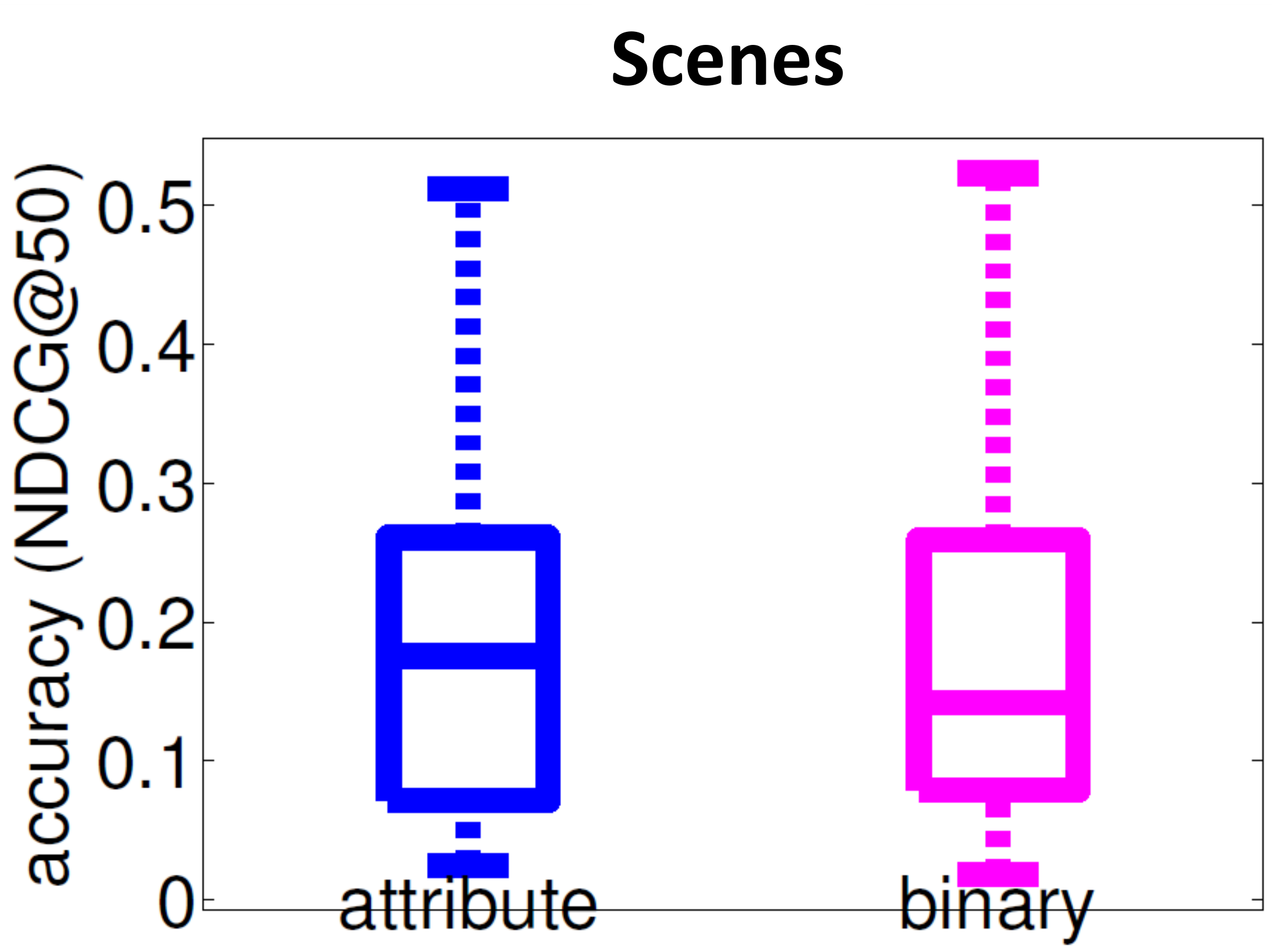}&
\includegraphics[width=0.22\textwidth]{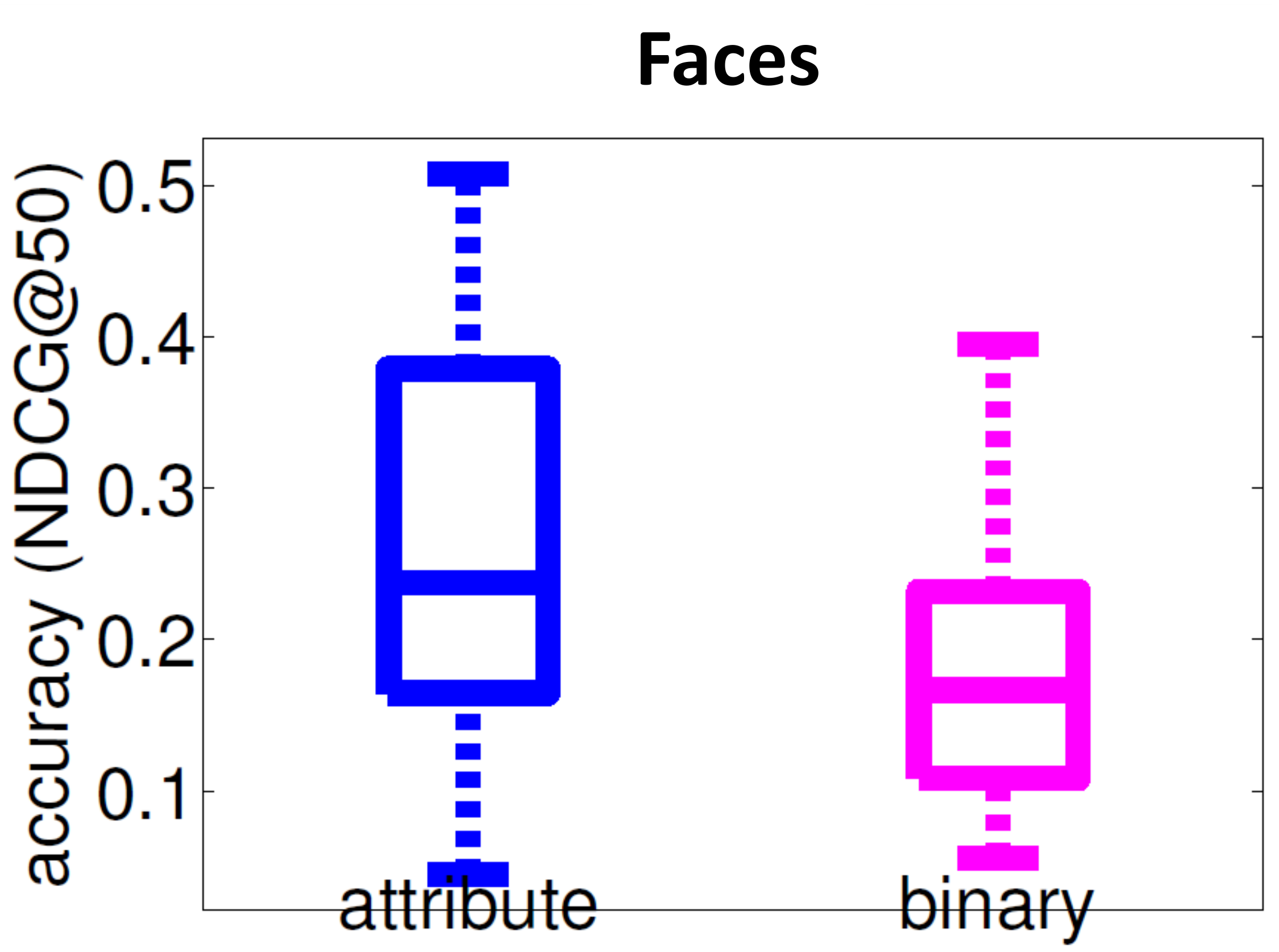}&
\includegraphics[width=0.22\textwidth]{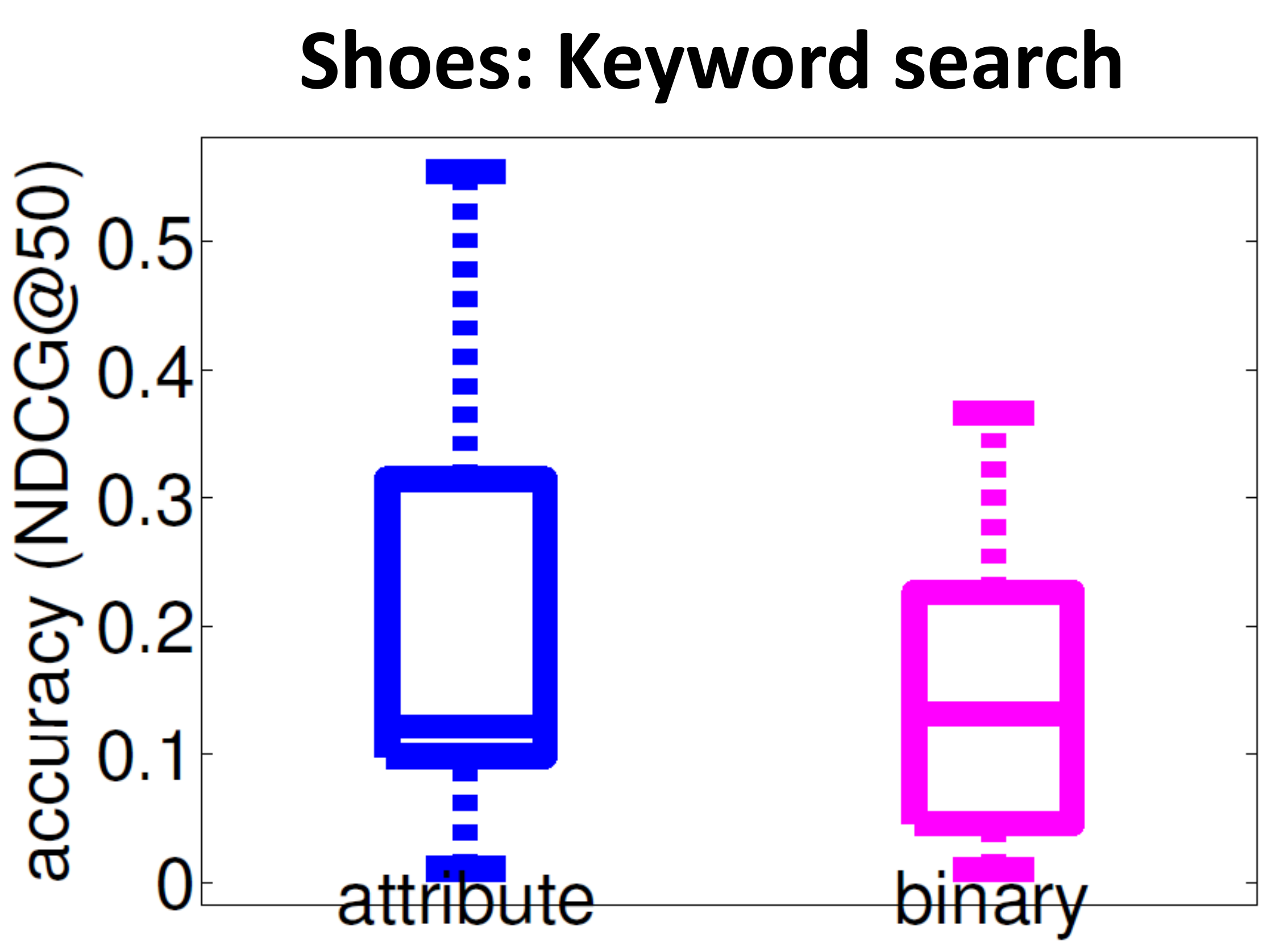}
\end{tabular}
\caption{Ranking accuracy with human-generated feedback with randomly chosen (first three plots) and keyword-initialized reference images (fourth plot).}
\label{fig:710}
\end{figure*}

\paragraph{Impact of reference images.}

\begin{table}[t]
\centering
\begin{tabular}{|c|c|c|c|c|}
\hline
Dataset-Method & Near & Far & Near+Far & Mid\\
\hline
Shoes-Attributes & .39 & .29  & \textbf{.40}  & .38\\
\hline
Shoes-Binary & .12  & .05  & \textbf{.27} & .06\\
\hline
Faces-Attributes & \textbf{.60} & .41  & .58  & .52 \\
\hline
Faces-Binary & .39  & .21  & \textbf{.64}  & .15 \\
\hline
Scenes-Attributes & \textbf{.53}  & .27 & .52  & .40 \\
\hline
Scenes-Binary & .18  & .18  & \textbf{.32}  & .11 \\
\hline
\end{tabular}
\caption{Ranking accuracy (NDCG@50 scores) as we vary the type of reference images available for feedback.  Bold values indicate the best performance in a row.}
\label{fig:type_prots}
\end{table}

The results thus far assume that the initial reference images are randomly selected, which is appropriate when the search cannot be initialized with keyword search.  We are interested in understanding the impact of the \emph{types} of reference images available for feedback.  Thus, we next control the pool of reference images to consist of one of four types: ``near'', meaning images close to the target image, ``far'', meaning images far from the target, ``near+far'', meaning a 50-50 mix of both, and ``mid'', meaning neither near nor far from the target.  Nearness is judged in the GIST/color feature space.

Table \ref{fig:type_prots} shows the resulting accuracies, for all types and all datasets using 100 queries and automatic feedback.  Both methods generally do well with ``near+far'' reference images, which makes sense.  For attributes, we expect useful feedback to entail statements about images that are similar to the target overall, but lack some attribute.  Meanwhile, for binary feedback, we expect useful feedback to contain a mix of good positives and negatives to train the classifier.  We further see that attribute feedback also does fairly well with only ``near'' reference images; intuitively, it may be difficult to meaningfully constrain precise attribute differences on an image much too dissimilar from the target.

\begin{figure*}[t]
\centering
\includegraphics[width=0.78\textwidth]{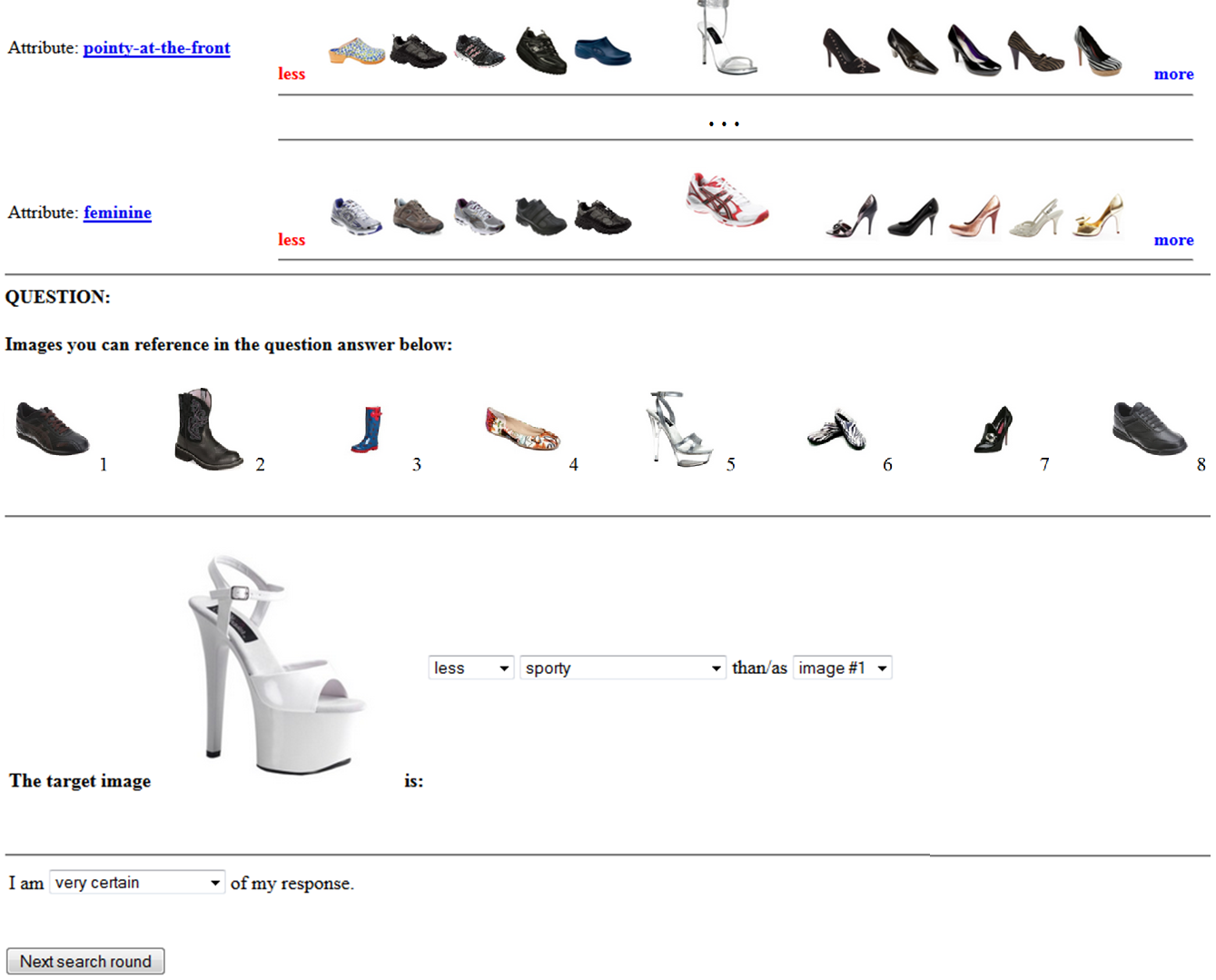}
\caption{The interface we use for the live user experiments for WhittleSearch.  The top rows illustrate the attribute meaning visually with examples.  Then the user is shown a series of eight reference images, and asked to compare a target image to a reference image of his choosing according to an attribute of his choosing using the drop-down boxes.  Finally, he must state his confidence in the response.}
\label{fig:interface_ws}
\end{figure*}

\paragraph{Ranking accuracy with human-given feedback.}

Having analyzed in detail the key performance aspects with automatically generated feedback, now we report results using human-generated feedback.  
Figure \ref{fig:interface_ws} shows the type of interface we used for these experiments.  At the top, we show users images from the bottom and top of our attribute rankers, in order to guide their answers and ameliorate the effect of the discrepancy between machine and user understanding of an attribute.
Figure~\ref{fig:710} (first three plots) shows the ranking correlation for both methods on 16 queries per dataset after one round of 8 feedback statements.  Attribute feedback largely outperforms binary feedback, and does similarly well on Scenes.  One possible reason for the scenes being less amenable to attribute feedback is that humans seem to have more confusion interpreting the attribute meanings (e.g., \emph{amount of perspective} on a scene is less intuitive than \emph{shininess} on shoes).

\begin{figure*}[htbp]
\centering
\includegraphics[width=1\linewidth]{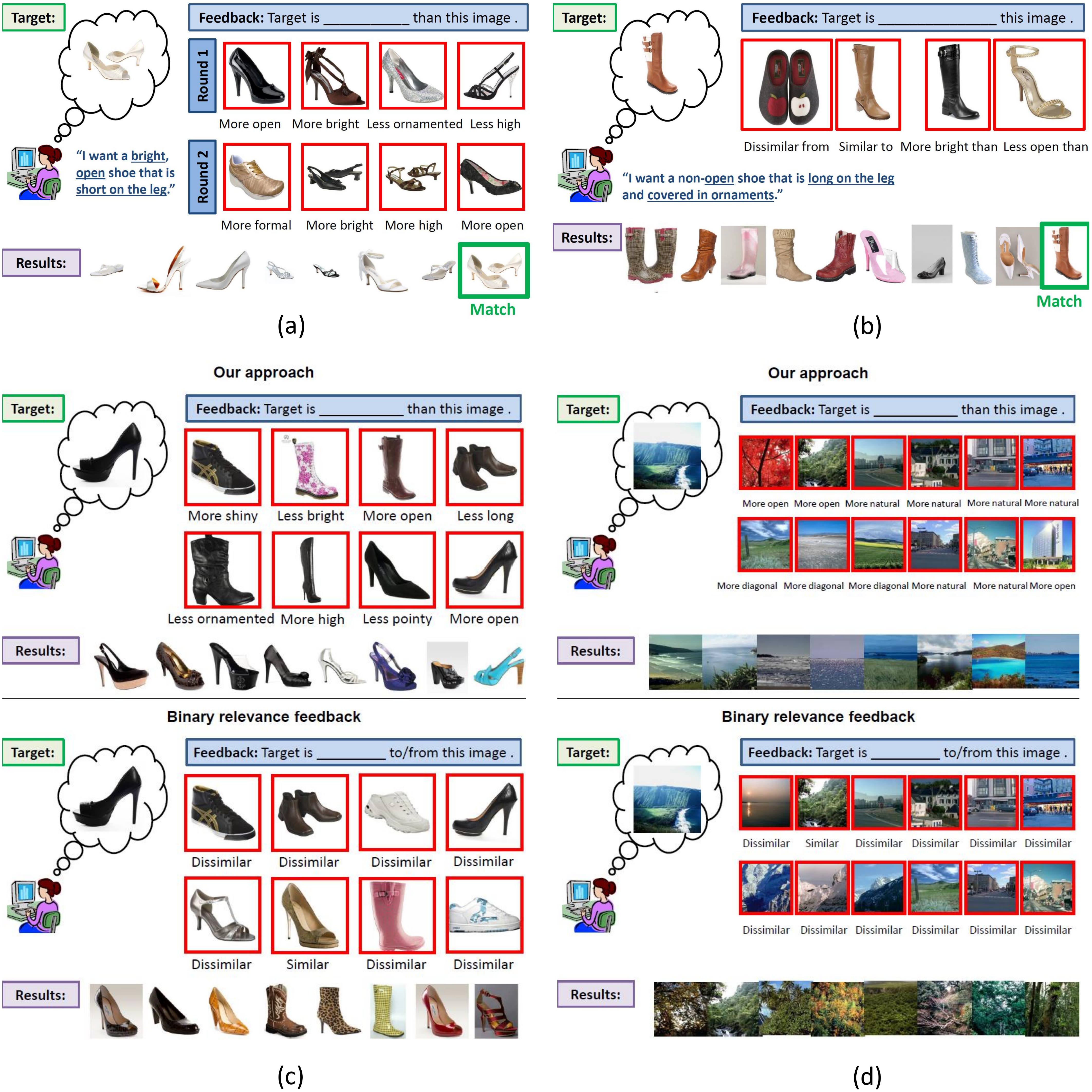}
\caption{(a) Example iterative search result with attribute feedback.  (b) Example search result with hybrid feedback.  (c, d) Example results for WhittleSearch (top) vs.~binary relevance feedback (bottom) on Shoes (c) and Scenes (d). For the Shoes example, while both methods retrieve high-heeled shoes, only our method retrieves images that are precisely as \emph{open} as the target image. This is because using the proposed approach, the user was able to comment explicitly on the desired openness property.  For the Scenes example, we show an interesting example of a target image that is hard to describe in words and likely has few very similar images in the database.  However, through our relative attribute constraints, we are able to retrieve better matches than the binary feedback baseline produces. A main issue for the baseline in this case is the lack of similar images among the reference images that the user can use to define positives.}
\label{fig:qual_big}
\end{figure*}

Next, we consider initialization with keyword search.  The Shoes dataset provides a good testbed, since an online shopper is likely to kick off his search with descriptive keywords.  Figure~\ref{fig:710} (fourth plot) shows the ranking accuracy results for 16 queries when we restrict the reference images to those matching a keyword query composed of three attribute terms.  Both methods get four feedback statements (we expect less total feedback to be sufficient for this setting, since the keywords already narrow the reference images to good exemplars).  Our method maintains its clear advantage over the binary baseline.  This result shows  (1) there is indeed room for refinement even after keyword search, and (2) the precision of attribute statements is beneficial.

\begin{figure}[t]
\centering
\includegraphics[width=1\linewidth]{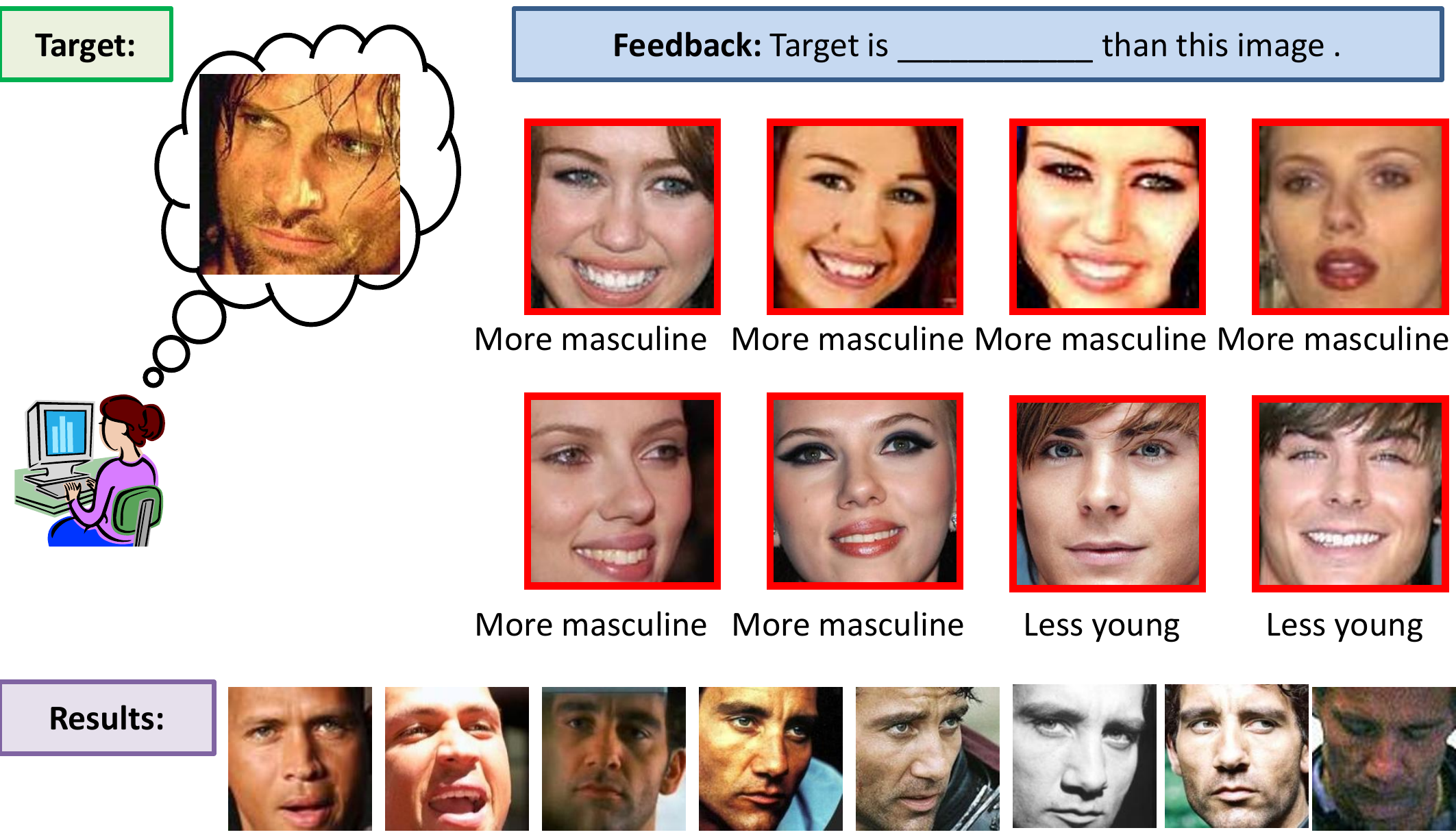}
\caption{A failure of our method. While the images our method retrieves do match the descriptions given by the user, in this case we fail to retrieve an image of the correct person. This failure may be due to the insufficiently rich description that the user provided.}
\label{fig:qual_failure}
\end{figure}

Figure~\ref{fig:qual_big} (a)
shows a real example search using relative feedback in WhittleSearch.  Note how the user's mental concept is quickly met by the returned images.  Furthermore, the user can comment very specifically on the heel height, by referring to both a very high-heeled shoe (in Round 1) and a shorter-heeled shoe (in Round 2).  
This example highlights the value of relative feedback: the user can precisely bound the range of acceptable strengths for each particular attribute.
In some cases, however, binary relevance feedback might be sufficient.  In Figure \ref{fig:qual_failure}, our method retrieves the correct images according to the user's descriptions.  But if the goal is to retrieve images of the person in the query, our method fails, while the binary relevance feedback method succeeds (not shown).  To combine the strengths of both approaches, we proposed a hybrid feedback approach in Section \ref{sec:hybrid}.
Figure~\ref{fig:qual_big} (b)
shows a real example using a hybrid of both binary and attribute feedback, as described in Section \ref{sec:hybrid}.  This suggests how a user can specify a mix of both forms of input, which are often complementary.

In Figure \ref{fig:qual_big} (c, d),
we present two real examples of search results for human-generated feedback with WhittleSearch, to compare our method qualitatively alongside the traditional binary relevance feedback approach.  
Each example shows one search iteration, where the 20 reference images are randomly selected (rather than ones that match a keyword search as the examples above) and annotated with constraints by users on MTurk.
For each result, the upper figure shows our method and the lower figure shows the binary feedback result for the corresponding target image.  
This figure shows the clear advantage of our relative attribute feedback approach over traditional binary feedback.  The user can retrieve more accurate results if he is allowed to compare the retrieved results to his target image for some particular visual property.

% ----- Supervision type -----

\begin{table}[t]
\centering
\begin{tabular}{|c|c|c|}
\hline
 & Class & Instance\\
\hline
Shoes & 26.10\% & \textbf{22.89\%}\\
\hline
Scenes & 38.92\% & \textbf{33.41\%}\\
\hline
Faces & \textbf{28.38\%} & 30.16\%\\
\hline
\end{tabular}
\caption{Errors for class-level vs.~image-level training.}
\label{fig:instance-vs-class}
\end{table}

\begin{figure}[t]
\includegraphics[width=1\linewidth]{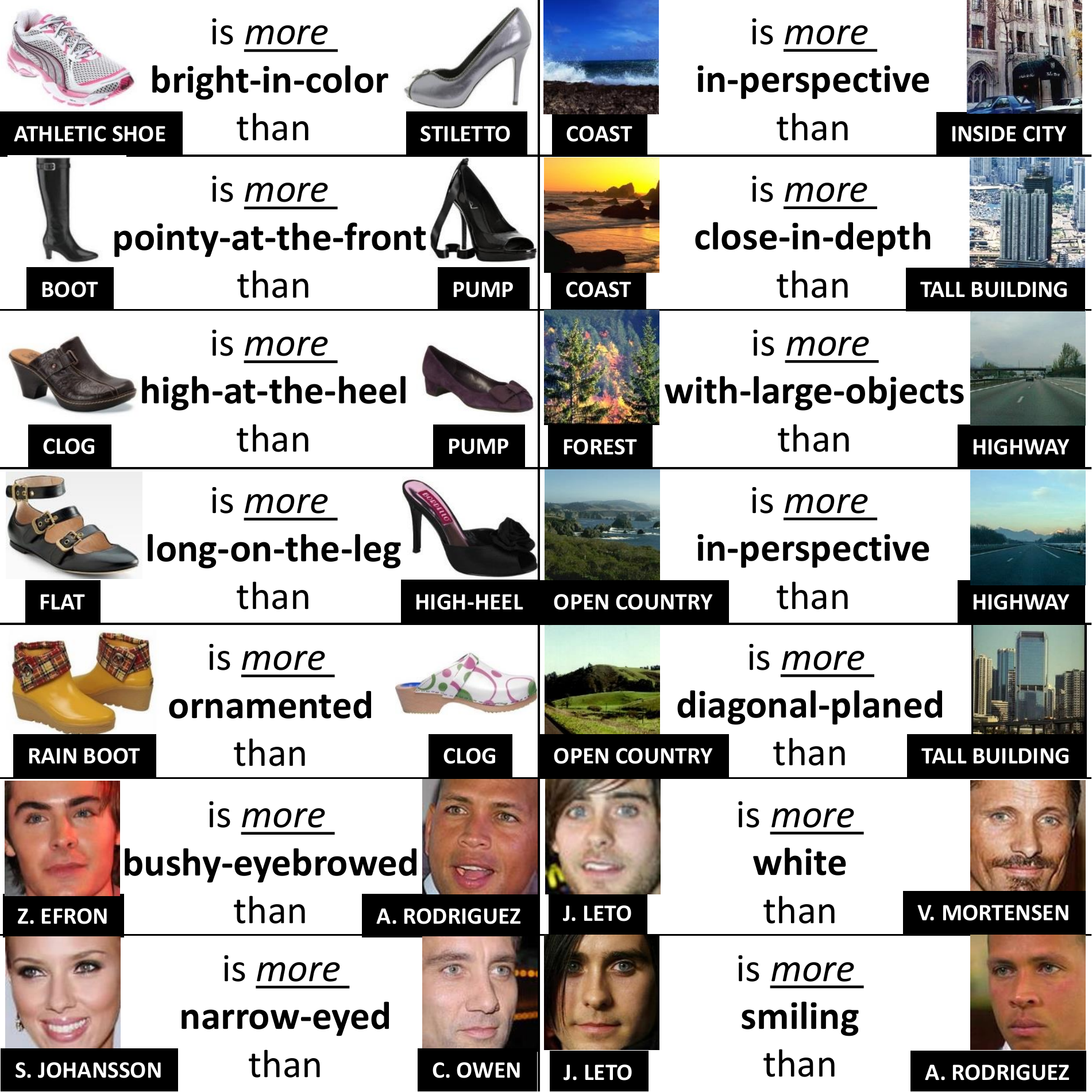}
\caption{Examples of dramatic disagreement between class-level and image-level annotations. For example, pumps are normally \emph{high at the heel} and clogs are \emph{flatter}, but the pump in the third row is lower at the heel than this particular clog. Inside-city images usually show a whole scene photographed down the street, but the inside-city scene in the top-right is the side of a building, and thus less \emph{in-perspective} than the coast image. Jared Leto is usually not \emph{smiling}, but in this particular picture (bottom-right) he is \emph{more smiling} than Alex Rodriguez.}
\label{fig:instance_examples}
\end{figure}

\paragraph{Consistency of relative supervision types.}

Next we examine the impact of how human judgments about relative attributes are collected to train the relative attribute models.  

For all results above, we train the relative attribute rankers using image-level judgments.  How well could we do if simply training with class-based supervision, i.e., ``coasts are \emph{more open} than forests''?  To find out, we use the relative ordering of classes given in~\cite{Parikh11b} for Faces and Scenes, and define them ourselves for Shoes (see Appendix).   We train ranking functions for each attribute using both modes of supervision.
 
Table \ref{fig:instance-vs-class} shows the percentage of $\sim$200 test image pair orderings that are violated by either approach.  Intuitively, instance-level supervision outperforms class-level supervision for Shoes and Scenes, where categories are more fluid.  
In additional experiments with 20 MTurk annotators, we find that the MTurkers' inter-subject disagreement on \linebreak instance-level responses was only 6\%, versus 13\% on category-level responses.  Both results support the proposed design for relative attribute training.
 
In Figure \ref{fig:instance_examples}, we show some examples where the instance-level ordering of two images with respect to some attribute differs from the ordering defined at the class-level.
We show annotations where  users had high confidence of these labels, and there was high inter-user agreement.

\subsection{Active WhittleSearch Results}
\label{sec:results-AWS}

We next test how well the active variant of our method guides the search process using attribute pivots, by comparing it to several alternative methods for interactive search.  Unless otherwise noted, we report results over 200 randomly chosen target images.

\paragraph{Baselines.}

We compare our Active WhittleSearch method, denoted \textsc{Active attribute pivots}, against the following six baselines: 

\begin{itemize}
\item \textsc{Attribute pivots} is a simplified version of our method that uses the attribute trees to select candidate images, but chooses randomly among the attributes in a round-robin fashion.
\vspace{0.1in}
\item \textsc{Active attribute exhaustive} uses entropy to select questions like our method, but it evaluates all possible $M$x$N$ candidate questions, where $M$ is the number of attributes and $N$ is the number of database images.
\vspace{0.1in}
\item \textsc{Top} selects the image that has the current highest probability of relevance and pairs it with a random attribute.  This method represents traditional interactive methods that assume an ``impatient" user for whom feedback exemplars and search results must be one and the same.  It is like the non-active version of WhittleSearch, except that it presents only \emph{one} reference image and allows only one statement to be given at each time. Unlike WhittleSearch, the user of the system cannot introduce variety in the feedback statements that are given, as he cannot exercise choice.
\vspace{0.1in}
\item \textsc{Passive} simply selects a random image paired with a random attribute for its question.
\vspace{0.1in}
\item \textsc{Active binary feedback} does not use statements about the relative attribute strength of images, but rather asks the user whether the exemplar is similar to the target.  This method uses a binary SVM to rank images, and treats similar images as positives and dissimilar images as negatives. It actively chooses the image whose decision value is closest to 0, as in~\citep{Tong01}.
\vspace{0.1in}
\item \textsc{Passive binary feedback} works as above, but randomly selects the images for feedback.
\end{itemize}

Note that the relative feedback methods all use the same relevance prediction function and only differ in the feedback they gather.  The tree-based methods stop asking questions about attribute $m$ once its leaf is reached or the user has given an ``equally" response for $m$.  All methods keep an image in consideration for feedback until all possible questions have been asked about it.

To thoroughly test the methods, we conduct both live experiments with real users as well as experiments where we simulate the user responses.
We generate the response for a question, ``Is the target image \emph{more, equally, or less} $m$ than $I_{p_m}$?'' using the difference in the predicted attribute values for the target $I_t$ and the pivot $I_{p_m}$.
For a response of ``equally'', we use a threshold derived from the training pairs of images labeled as similar with respect to $m$.
Note that this protocol is in line with standard validation for active learning, where the algorithm receives the labels for those examples it queries, even if a human is not answering ``live'' in the loop.
The predicted attribute values are an extrapolation of the ground-truth labels we have obtained from users. 
We initialize all attribute search methods with the same feedback constraint.

For binary relevance feedback, we respond with ``similar'' if the target and exemplar images are within one standard deviation of the learned distances used for the ground truth ranking.  We initialize the baseline with one positive and one negative image by peeking at the distances between the target image and a pool of 40 images, and selecting the closest image as a positive and the furthest as a negative.  This simulates a user starting the search with feedback on a page of random images.  If anything, it is generous to the baseline, since our method gets only one ``bit'' of feedback at the onset, while the binary feedback baselines get two.

We again add Gaussian noise to both the relative attribute feedback and binary feedback methods in order to account for the discrepancy between perceived and predicted attributes and appearance.

\paragraph{Comparison of likelihood models.}

\begin{figure}[t]
\includegraphics[width=1\linewidth]{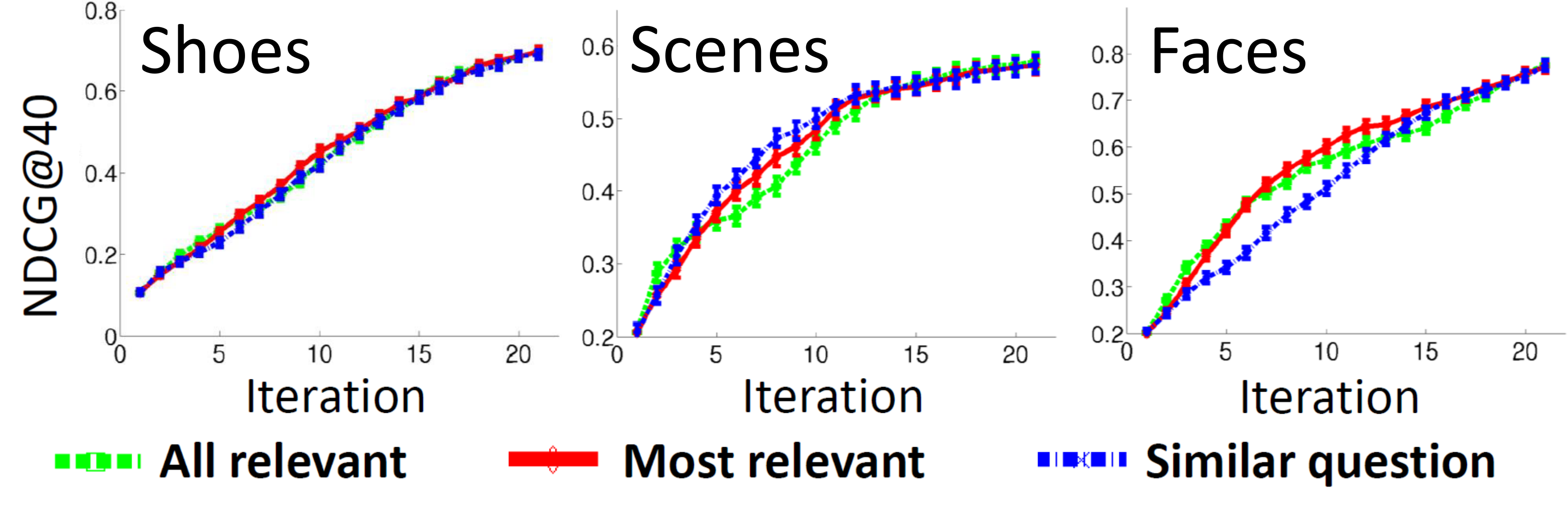}
\caption{Comparison of the proposed models for the likelihood of a user's response.  Best viewed in color.}
\label{fig:likelihood}
\end{figure}

Figure \ref{fig:likelihood} compares the three proposed methods of predicting the user response. \textsc{Most Relevant} consistently performs well on all datasets, and outperforms the other two methods on all but the Scenes.  This suggests that our best guess at the target tends to be a sufficient proxy, having a fairly similar attribute signature.  \textsc{All Relevant} performs similarly but is slightly weaker, indicating that isolating the most relevant instance gives a ``cleaner'' likelihood than attempting to refine it with our uncertainty about each relevant instance.  \textsc{Similar Question} performs the best for a fraction of the iterations on Scenes, but does poorly on Faces. This is likely because we cannot estimate attribute similarity reliably due to the distinct face attributes (e.g., face \emph{chubbiness} has no strongly correlated attributes, whereas scene \emph{openness} does).  In all remaining results, we use the \textsc{Most Relevant} method.

\begin{figure}[t]
\includegraphics[width=1\linewidth]{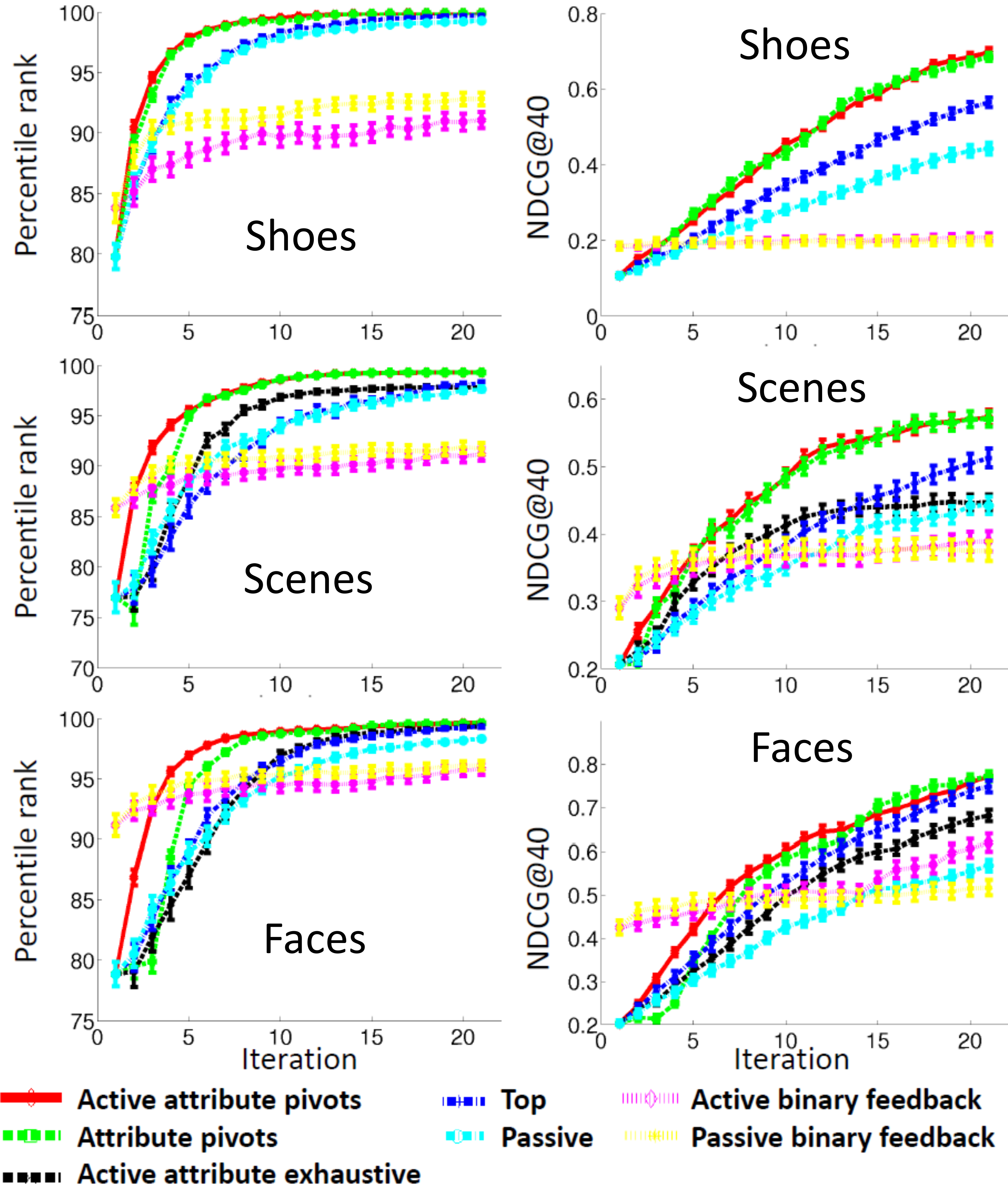}
\caption{Comparison of Active WhittleSearch to alternative interactive search methods on the three datasets.  For both metrics, higher curves are better.  Best viewed in color.}
\label{fig:baselines}
\end{figure}

\begin{figure*}[t]
\centering
\includegraphics[width=0.78\textwidth]{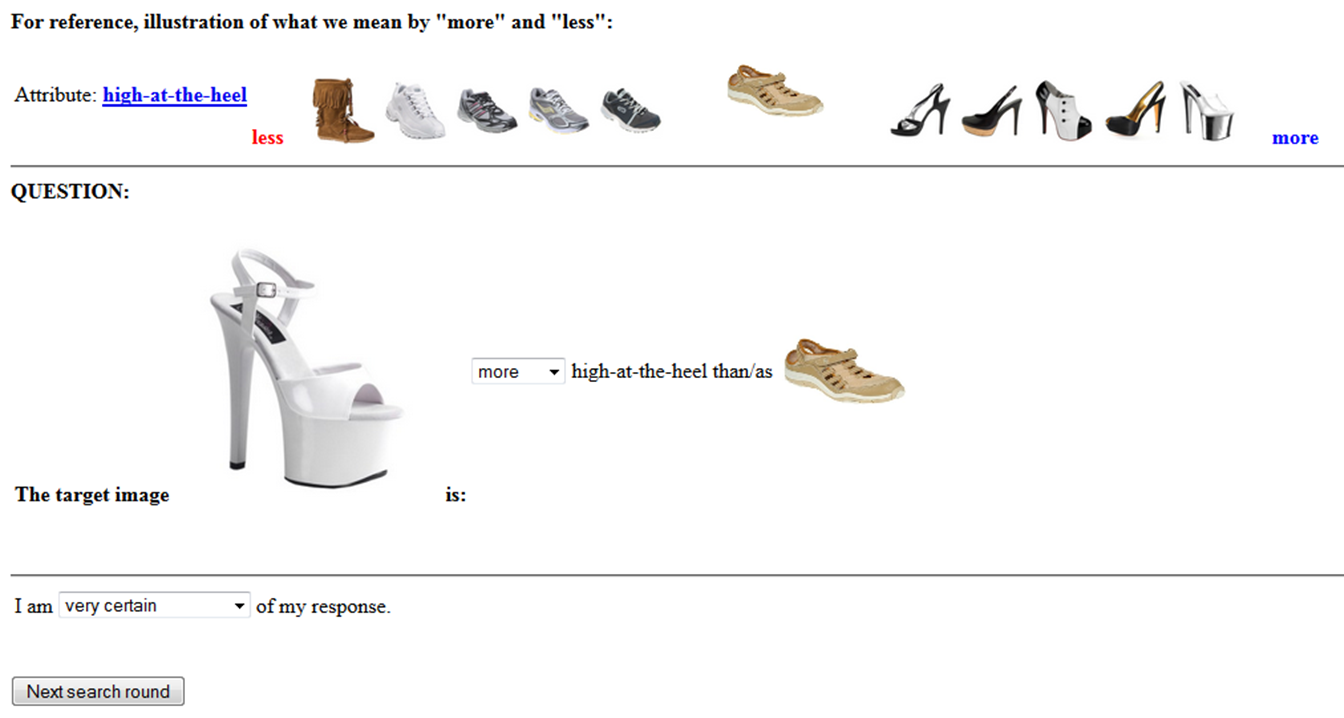}
\caption{The interface we use for the live user experiments for Active WhittleSearch.  The top row illustrates the meaning of the system-selected attribute for this round of feedback.  Then the user is asked to compare the displayed target image to the selected reference image according to that attribute, by selecting ``more'', ``less'', or ``equally''.  Finally, he must state his confidence in the response.}
\label{fig:interface_aws}
\end{figure*}

\paragraph{Comparison to existing methods.}

Figure~\ref{fig:baselines} compares our method to the six baselines on all three datasets.  
Overall, our method finds the target image most efficiently.
We see that our full active approach outperforms the round-robin variant of our method (\textsc{Attribute pivots}), with an average percentile rank 7.6\% better after only 3 iterations.  This shows actively interleaving the trees allows us to focus on attributes that better distinguish the relevant images.

Our method is also more effective than \textsc{Active attribute exhaustive}.\footnote{The exhaustive baseline was too expensive to run on all 14K Shoes.  On a 1000-image subset, it does similarly as on other datasets.} This shows that the binary tree structures serve as a form of regularization, helping our method focus on those questions that \emph{a priori} may be most informative to ask.  
Intuitively, if a user has ruled out a subtree (``The target image is bluer than the reference image with blueness X.''), it is likely redundant (low information gain) to ask how the target compares to more data on that path (``Is the target image bluer than this other reference image with blueness X - Y?''), i.e., to ask the user to comment on something even less blue than the previous exemplar.
The exhaustive method might be more prone to selecting outliers which are not actually informative, due to potential noise in the active selection which arises out of the need to estimate the likelihood of different user responses.  In contrast, our method picks pivots which, even if there are small errors in the entropy estimation, will be informative as they split the search space in half.
Furthermore, our method is orders of magnitude faster (see Table~\ref{fig:times}).

\begin{table}[t]
\centering
\begin{tabular}{|c|c|c|c|}
\hline
Method/Dataset & Shoes & Scenes & Faces\\
\hline
Active attribute pivots (Ours) & 0.05 & 0.01 & 0.01 \\
Active attribute exhaustive & 656.27 & 28.20 & 3.42 \\
\hline
\end{tabular}
\caption{Selection time for one iteration of our method vs.~the exhaustive active baseline, in seconds.}
\label{fig:times}
\end{table}

The results in Figure~\ref{fig:baselines} also show the striking advantage of relative attribute feedback compared to binary relevance feedback, as we also demonstrated in the previous section.  Binary feedback has an advantage in the first few iterations, likely because we generously initialize it with two feedback statements.  However, the relative attribute methods quickly surpass binary feedback.  We find that both feedback modes require similar user time: 6.4 s for relative, and 5.5 s for binary, and so the trends remain if we plot rank as a function of user time.  Interestingly, we find that \textsc{Passive binary feedback} is actually stronger than its active counterpart for this data.  This is likely because images near the decision boundary are often similar (and negative), whereas the passive approach samples more diverse instances (and hence gets more positives).  

Finally, we outperform \textsc{Top}, showing that relative attribute feedback alone need not offer the most efficient search.
Rather, it is important to give comparative constraints on well-chosen images.

In practical terms, we are interested in how many iterations it takes to get the target in the top 40 most relevant images, since that is how many images fit on a typical search page (e.g., on Google).  On average our method uses 12, 10, and 4 iterations to place the target in the top 40 for Shoes, Scenes, and Faces, vs. 21, 21, and 9 iterations for \textsc{Top}.  Thus, our method saves a user up to 70 seconds per query.

We also tested a method that does a hard pruning of images on the irrelevant branches of an attribute tree, as dictated by user feedback.  It incorrectly eliminates the true target for about 93\% of the queries, clearly supporting the proposed probabilistic formulation.

\paragraph{Results with live users.}

Next, we test our method ``live" in real time with Mechanical Turk workers, using an interface similar to the one shown in Figure \ref{fig:interface_aws}.  
We compare its performance against the two strongest baselines, \textsc{Attribute pivots} and \textsc{Top}.   
The workers answer a series of five questions that each of the three methods pose about the same target image.
We issue 50 queries for Shoes-1k (a random 1000-image subset of Shoes), Scenes, and Faces-Unique (a set of one image for each of 200 individuals from the original PubFig dataset \citep{Kumar09}, using the six most reliably predictable attributes).  
We eliminate any queries where one or more methods did not receive five complete feedback iterations.
All methods share one simulated feedback statement at iteration 0, which we do not plot.
We stop updating the probabilities of relevance for a method once this method places the target image in the top 40 images.

\begin{figure}[t]
\includegraphics[width=1\linewidth]{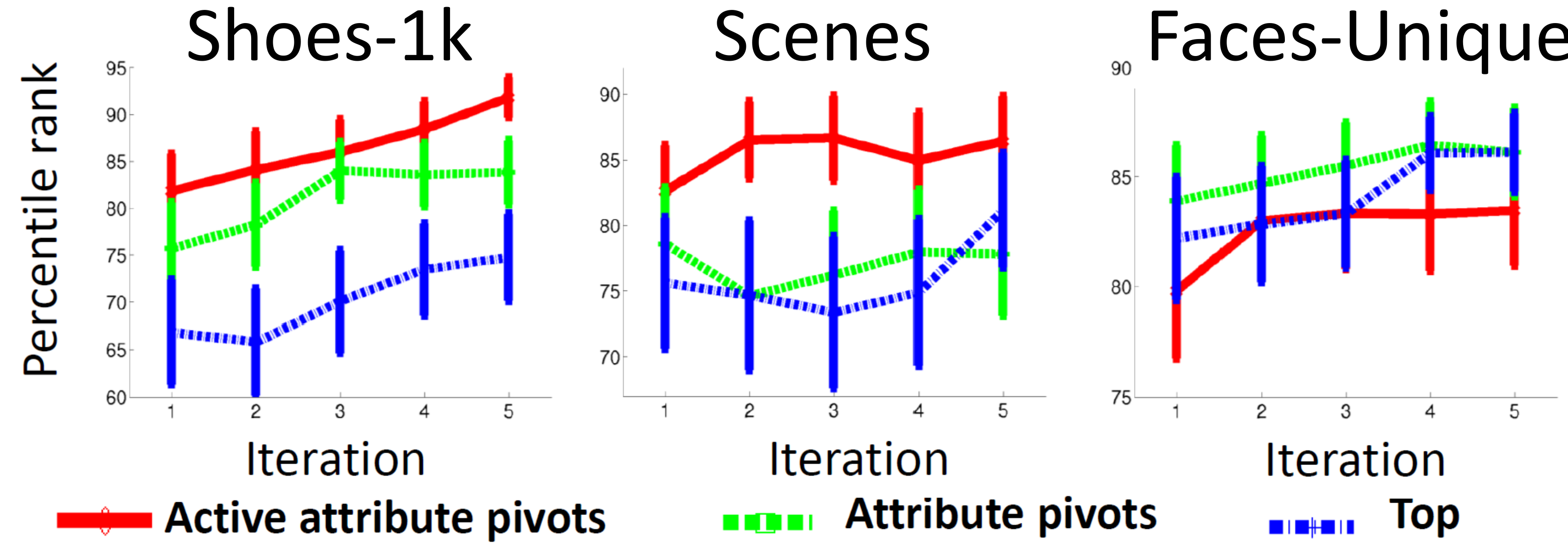}
\caption{Our Active WhittleSearch method makes quick and reliable choices, allowing the MTurk users to more efficiently find the target.}
\label{fig:live}
\vspace{-0.2in}
\end{figure}

\begin{figure*}[t]
\centering
\begin{tabular}{cc}
\subfigure[Shoes-1k]{
\includegraphics[width=0.45\textwidth]{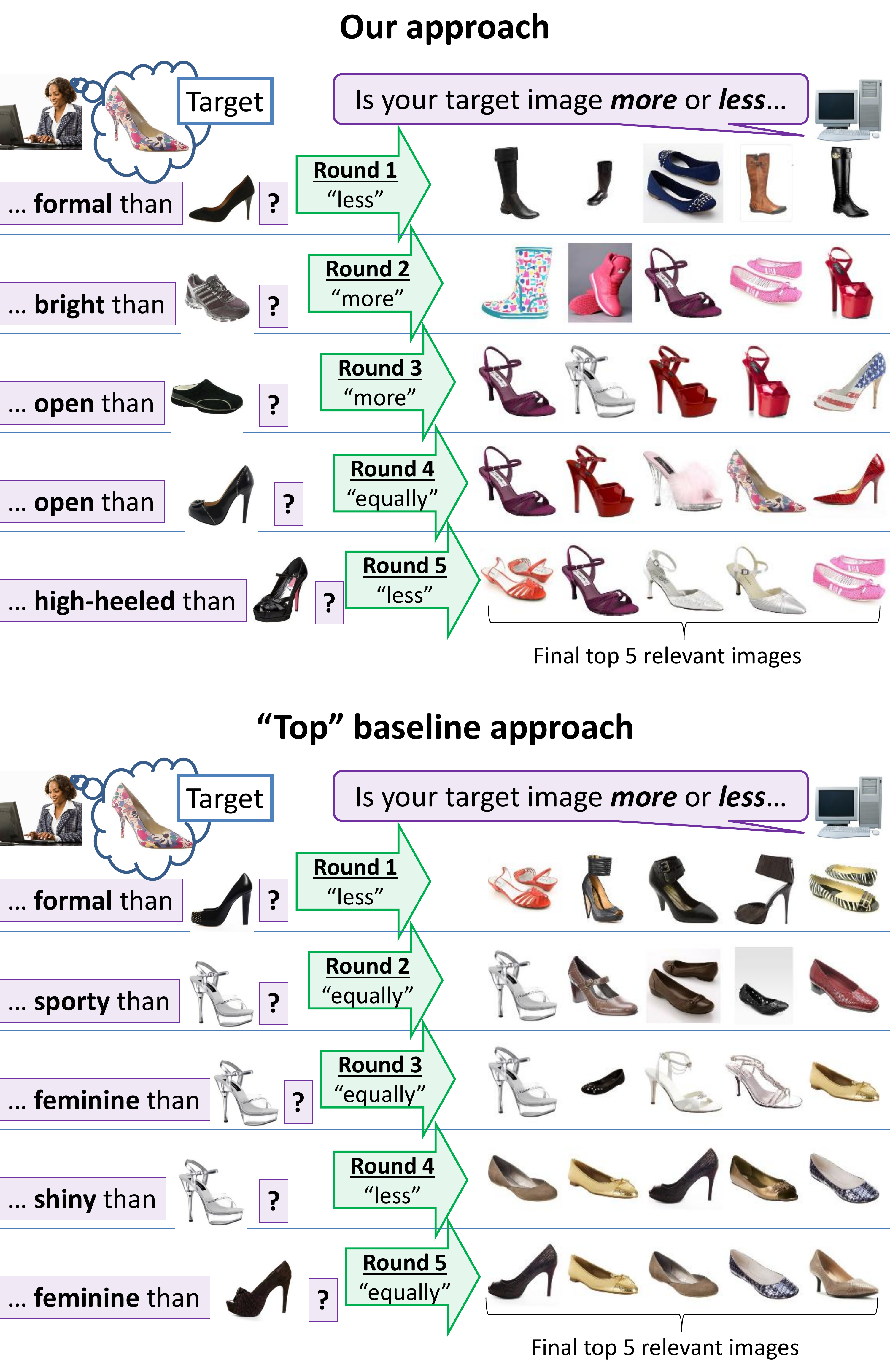}}
\hspace{0.3in}
\subfigure[Scenes]{
\includegraphics[width=0.45\textwidth]{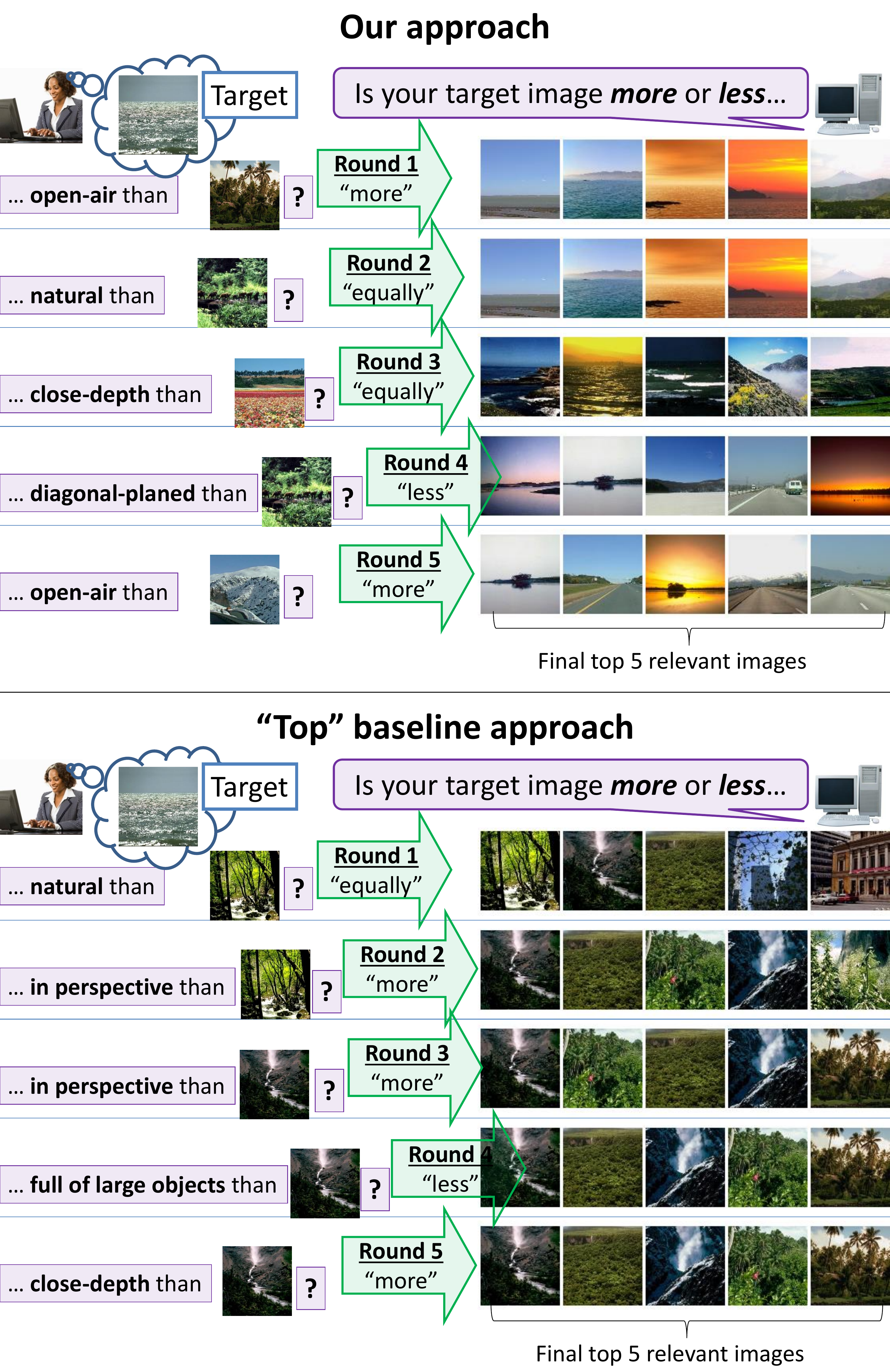}}
\end{tabular}
\caption{Example live user search results comparing our method (top) to the \textsc{Top} baseline (bottom) on Shoes-1k (a) and Scenes (b). Using the user's feedback on the left, we retrieve the images on the right at the top of the results list.  See text for details.}
\label{fig:qual_aws_more}
\end{figure*}

In order to get richer feedback from users, we allow users to express their confidence in their responses, and give twice the weight to constraints for which the user says ``a lot more (less)'' when computing the relevance probabilities. 

Note that this live experiment is only possible because our method can make decisions in real time, unlike the exhaustive active learning method.

Figure \ref{fig:live} shows the results.   Consistent with our simulated user results above, we see that typically our method ranks the target image better than the baselines do.  We find this a very encouraging result, given the noise inherent in MTurk responses (in spite of our best efforts at qualification tests) and the difficulty of predicting all attributes reliably.  Our informativeness predictions on Faces-Unique are imprecise since the facial attributes are difficult for both the system and humans to compare reliably (e.g., it is hard to say who among two white people is \emph{whiter}). This difficulty seems to hurt all methods, judging by their flatter curves.
Since the rank metric does not give any credit for finding an image very close to the target, we also asked a separate set of workers to judge whether any of the top 10 ranked images were ``very similar" to the target.  For Shoes-1k, our method takes only 1.9 iterations on average to find one that is very similar, whereas \textsc{Attribute pivots} requires 2.4 and \textsc{Top} requires 3.15.

Figure \ref{fig:qual_aws_more} presents examples of real live searches done by workers on Mechanical Turk with the Active WhittleSearch system.  We show how our method and \textsc{Top} rank images (shown on the right-hand side) based on the supplied user feedback (shown on the left-hand side).  In each figure, our method is shown on top, followed by \textsc{Top} underneath. For simplicity, we show ``a lot more/less'' responses as simply ``more/less''.  In Figure \ref{fig:qual_aws_more}(a), we see how our method quickly converges on shoes that look like the target (\emph{bright high-heeled pointy} shoes). 
Our method asks questions that are crucial in describing the shoe precisely (it is a \emph{high-heeled} but not a \emph{formal} shoe, and it is \emph{more open} than other high-heeled shoes). In contrast, \textsc{Top} gets stuck asking questions about the same shoe, and moreover, asking questions whose answers might be redundant (i.e., about \emph{sportiness} and its near-opposite \emph{femininity}).  In Figure \ref{fig:qual_aws_more}(b), 
our method asks about properties that are important for distinguishing the target image from other images, namely open-air. 
Only our method is able to provide acceptable top results.

\subsection{Comparing WhittleSearch and Active WhittleSearch}
\label{sec:results-comparison}

So far, we have demonstrated the advantages of relative attribute feedback, as well as the benefit of actively selecting the images shown for such relative attribute feedback.  We have also discussed the conceptual advantages of the user-guided version of WhittleSearch and its system-guided active selection version, in Section \ref{sec:discussion-WS-vs-AWS}.  

\begin{figure*}[t]
\includegraphics[width=1\linewidth]{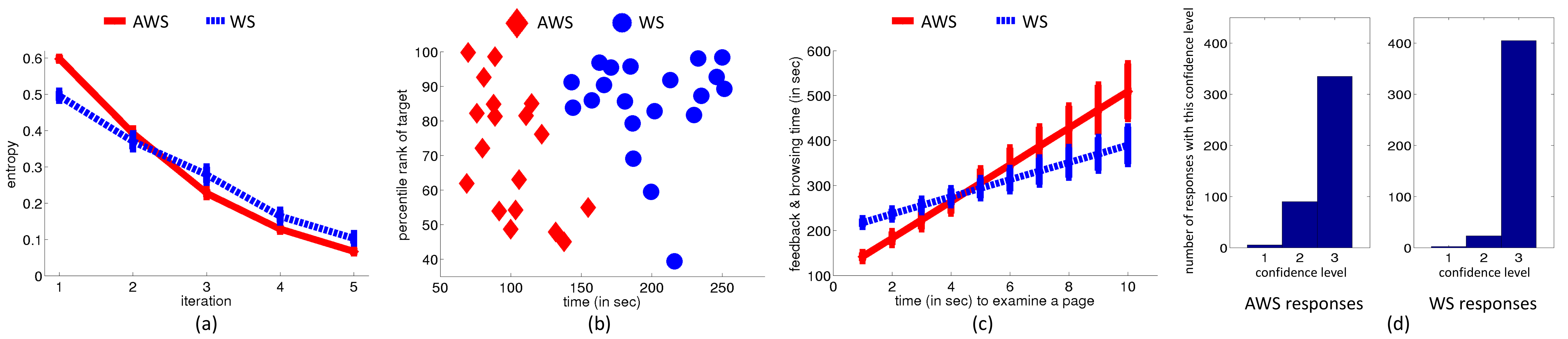}
\caption{Comparison of WhittleSearch (WS) and Active WhittleSearch (AWS). (a) System entropy for WhittleSearch (WS) and Active WhittleSearch (AWS) (lower is better). (b) Percentile rank of target vs. time required for feedback (higher rank and lower time are better). (c) Total time, with rank converted to time (see text). (d) Confidence of human responses for WhittleSearch (WS) and Active WhittleSearch (AWS).}
\label{fig:comparison}
\end{figure*}

Next, we compare the two versions of our method experimentally, using the Shoes dataset.  We conduct experiments where users provide one feedback statement at each of five iterations, whether that is chosen by the user 
from among those that are ranked highest at the previous iteration (for WhittleSearch), or actively chosen by the system (for Active WhittleSearch).  Each of 20 queries is submitted to five workers, and each worker completes the task for the same query for both methods.  
We time the user responses at each iteration.  We manually remove outliers in terms of time, and queries for which the users provided obviously incorrect responses, for both methods.  

In Figure \ref{fig:comparison} (a), we demonstrate that Active WhittleSearch does indeed reduce the overall entropy of the system better than WhittleSearch, which is the objective that Active WhittleSearch uses when selecting comparisons for feedback.  We plot how entropy decreases as the system receives more feedback over five iterations.  
The entropy estimates in the first few iterations are inaccurate due to the system having received too little feedback to estimate relevance accurately.  This likely explains why Active WhittleSearch is initially weaker at reducing entropy, but after two iterations, it starts to reduce entropy faster than WhittleSearch, thus achieving its main objective.

Next, we examine how entropy reduction affects the actual user experience, as measured by the success of search results as a function of the amount of feedback effort.  
In Figure \ref{fig:comparison} (b), we plot the median final percentile rank of the target image per query, and the median total time it took to provide all feedback statements for that method.  The time for feedback captures the time that users spend to examine the reference images and attribute vocabulary and consider the possible combinations thereof they can use for a feedback statement, as well as the time they spend actually submitting the selected feedback.  
If no options are given and the system simply presents the human user with a single question, then the time for feedback simply involves deciding on the answer to that question (i.e., ``more'', ``less'', or ``equally'').   Since WhittleSearch gives the user more freedom and the user needs to examine options and select among them, that version requires more time for feedback than the active version, which could potentially be a disadvantage to an impatient user.
That said, WhittleSearch often achieves high accuracy rates as a payoff for the user time invested.

To better depict how the user effort and quality of results are tied together, we next devise a unified metric for evaluation; see Figure \ref{fig:comparison} (c).  This metric  measures both how long it takes to provide a specific form of feedback, and how effectively this feedback enables the system to retrieve results, captured by the rank of the target image.  In particular, we sum the time for providing the feedback and the time required to examine the results.  The latter term corresponds to the rank of the target image converted to time, using a varying number of seconds that are required to examine a page of 40 images.  In other words, if the target image is shown at rank 70, it will be on page two of the search results, and if it takes 4 seconds to examine a page, the total time to examine the results will be 8 seconds.  We plot results as a function of the time to examine a page because examining a page of results can take a short amount of time---if the target image has very prominent and easy to spot distinctive features or if all of the results are obviously very different than the target image---or more time---if some of the results are similar to the target and the user needs to look more carefully to determine if there is an actual match.  We find that perusing a page of 40 image results takes 5.7 seconds on average, hence the choice of range we use on the x-axis of Figure \ref{fig:comparison} (c).

In Figure~\ref{fig:comparison}(b), we see that Active WhittleSearch is \linebreak cheaper in terms of user time, but achieves slightly worse ranks for the target image.  Because WhittleSearch achieves better ranks than Active WhittleSearch on average but is slower to use, the user-guided version outperforms the system-guided one when the cost of examining a page of results starts to dominate the cost of providing feedback, as seen in Figure~\ref{fig:comparison}(c). 
This result illustrates how different versions of the WhittleSearch system might be preferable in different contexts and for different tasks.

\begin{figure}[t]
\includegraphics[width=1\linewidth]{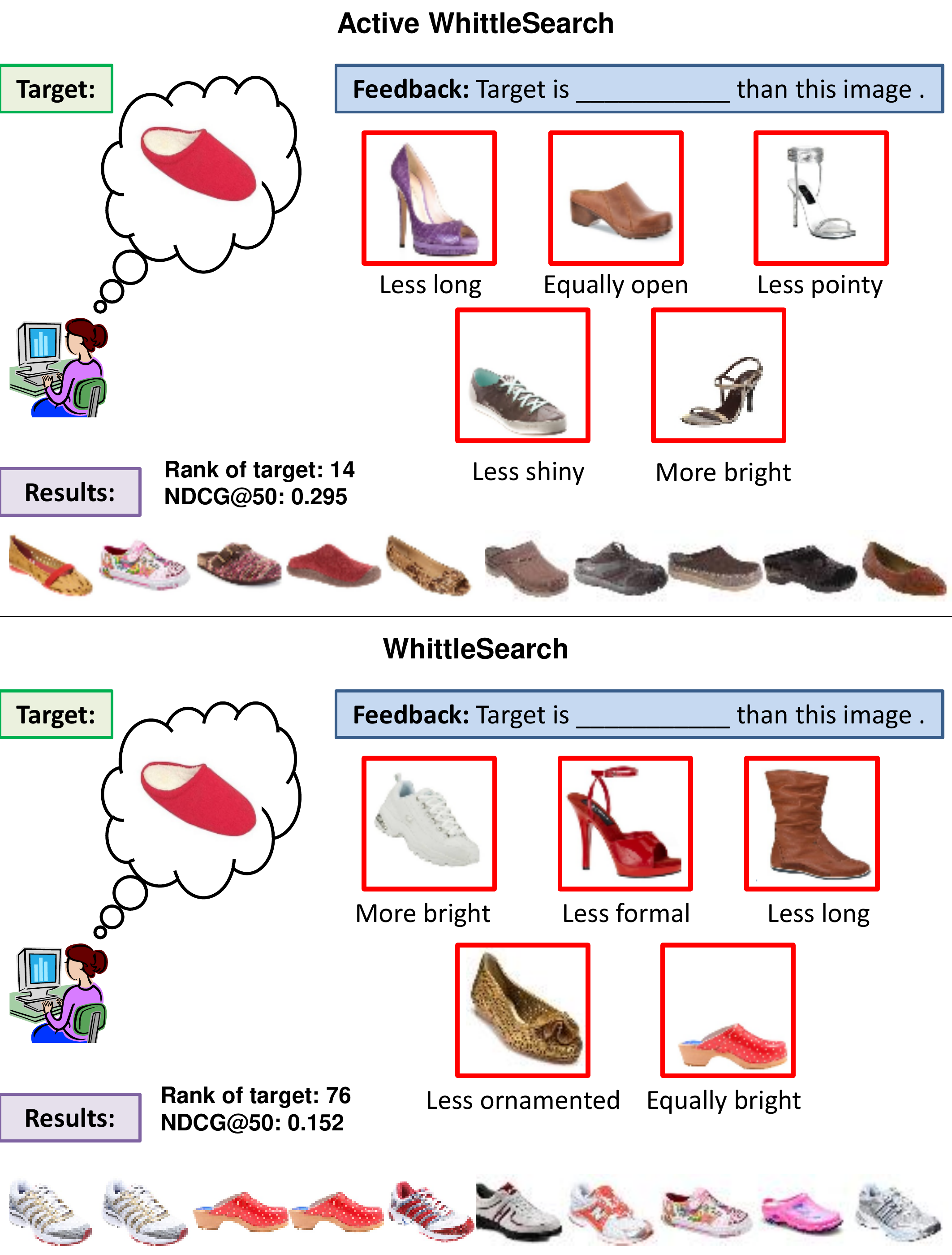}
\caption{A case where Active WhittleSearch is most useful.  Observe the discriminative questions selected by the active system---not only in terms of attributes like \emph{bright-in-color} and \emph{long-on-the-leg}, but also in terms of the images involved in the comparison along those attribute dimensions.  For example, the user of WhittleSearch chooses to comment on the relevant \emph{long-on-the-leg} property, but there are a lot more images that are \emph{less long-on-the-leg} than a boot (bottom), compared to those that are \emph{less long-on-the-leg} than a pump (top).}
\label{fig:AWS_conf_best}
\end{figure}

To examine possible reasons for the performance of the two versions of the system, in Figure \ref{fig:comparison} (d) we show a histogram of the confidences that users reported for their responses.  We plot the average certainty that the user provided over the five iterations, with 3 being most certain and 1 being uncertain. We see that human responses on WhittleSearch are much more certain than those for its system-guided counterpart, likely because users often comment on the most obvious relationships of target and reference images when they are given a choice. This explains Active WhittleSearch's inferior performance in terms of rank, in Figure \ref{fig:comparison} (b).  However, we observe that when all five MTurkers agree on all of the Active WhittleSearch responses, which occurred for one query, Active WhittleSearch is better.   Figure \ref{fig:AWS_conf_best} shows this example for one of the five users.  This is encouraging because it indicates that if we can pick feedback requests that are informative \emph{and also likely to be answered with confidence}, our active approach can produce even more accurate search results.
Thus, a natural direction for future work is to incorporate a user-confidence model into the system.

% -----

\section{Conclusion}

We proposed an effective new form of feedback for image search using relative attributes. 
In contrast to traditional binary feedback, our approach allows the user to precisely indicate how the results compare with his mental model. 
Building on this idea, we develop a system-guided version of the method which actively engages the user in a \emph{relative} 20-questions-like game, where the answers are visual comparisons. 
Compared to existing active and passive methods, our pivot-based formulation is both more efficient (by orders of magnitude) and more accurate in practice. 

In-depth experiments with three diverse datasets show relative attribute feedback's clear promise, and suggest interesting new directions for integrating multiple forms of feedback for image search. 
Results demonstrate that our system-guided approach can rapidly pinpoint the visual target using a series of well-chosen comparative queries. 

In future work, we plan to explore ways to more fully model uncertainty in the search system.  This can include, for example, representing the user's confidence when computing our active selection criteria, or accounting for the confidence of the attribute models themselves.  
Furthermore, we would like to encourage diversity in the questions we ask the user, incorporate strategies for ensuring that the questions we ask are not too difficult, and develop an approach where control can be adaptively transferred between the user and the system.
We will study ways to efficiently learn a new attribute on the fly, to allow the user to define new attributes when the current vocabulary is no longer useful.  We are also interested in developing ways to allow for more exploration during search, and for assignment of different weights to feedback on different attributes.

\begin{acknowledgements}
%If you'd like to thank anyone, place your comments here
%and remove the percent signs.
We thank the anonymous reviewers for their helpful feedback and suggestions. This research was supported by ONR YIP award N00014-12-1-0754 (K.G. and A.K.) and Google Faculty Research Award (D.P.).
\end{acknowledgements}

% BibTeX users please use one of
%\bibliographystyle{spbasic}      % basic style, author-year citations
%\bibliographystyle{spmpsci}      % mathematics and physical sciences
%\bibliographystyle{spphys}       % APS-like style for physics
%\bibliography{}   % name your BibTeX data base

\bibliographystyle{spbasic}
\bibliography{refs}

\clearpage
\onecolumn

\textbf{Appendix A}

\begin{table}[h]
\centering
\begin{tabular}{|c|c|c|c|c|c|c|c|c|c|c|}
\hline
Attribute/Class & Athletic & Boots & Clogs & Flats & Heels & Pumps & Rain Boots & Sneakers & Stiletto & Wedding \\
\hline
Pointy at the front  &   2 &    6  &   3  &   5  &  10  &   9 &    4  &   1  &   8  &   7\\
\hline
Open &    3  &   2   &  8  &   5   &  7   &  6  &   1  &   4   &  9  &  10\\
\hline
Bright in color   &  6 &    1 &    2 &    8 &    4  &   3  &  10  &   7  &   9 &    5\\
\hline
Covered w/ ornaments   &  4  &   9 &    6 &    5  &   8  &   7  &   1 &    3  &  10  &   2\\
\hline
Shiny &    2 &    9  &   4  &   3  &   6 &    5 &    8 &    1 &   10 &    7\\
\hline
High at the heel &    4   &  6  &   5   &  1   &  9  &   8  &   3   &  2   & 10  &   7\\
\hline
Long on the leg &    7 &    9  &   2   &  3   &  6  &   5  &  10   &  8 &    4 &    1\\
\hline
Formal  &   3  &   6  &   4  &   7    & 9   &  8 &    1    & 2 &     5  &  10\\
\hline
Sporty  &  10 &    5  &   6  &   7   &  4   &  3  &   8  &   9  &   1  &   2\\
\hline
Feminine &    1 &    6 &    4 &    5  &  10  &   9    & 3 &    2  &   8  &   7\\
\hline
\end{tabular}
\caption{Ordering of classes for the attributes in the Shoes dataset. A score of 10 denotes that this class has the attribute the most, and 1 denotes the class has it the least.}
\label{fig:shoes_class_orders}
\end{table}

%% Non-BibTeX users please use
%\begin{thebibliography}{}
%%
%% and use \bibitem to create references. Consult the Instructions
%% for authors for reference list style.
%%
%\bibitem{RefJ}
%% Format for Journal Reference
%Author, Article title, Journal, Volume, page numbers (year)
%% Format for books
%\bibitem{RefB}
%Author, Book title, page numbers. Publisher, place (year)
%% etc
%\end{thebibliography}

\end{document}